\documentclass[10pt,twocolumn,letterpaper]{article}

\usepackage{cvpr}              
\usepackage{url}
\usepackage{multirow}
\usepackage{newfloat}
\usepackage{listings}
\usepackage[utf8]{inputenc} 
\usepackage[T1]{fontenc}    
\usepackage{url}            
\usepackage{booktabs}       
\usepackage{amsfonts}       
\usepackage{amsthm}         
\usepackage{amsmath}        
\usepackage{nicefrac}       
\usepackage{microtype}      
\usepackage[table]{xcolor}
\usepackage{mathtools}
\usepackage{amsthm}
\usepackage{wrapfig}
\usepackage[most]{tcolorbox}
\usepackage{algorithm}
\usepackage{algorithmicx}
\usepackage{algpseudocode}
\usepackage{fontawesome5}
\usepackage{cuted}
\usepackage{colortbl}

\usepackage{amsthm}

\usepackage{graphicx} 
\usepackage{float}
\usepackage{enumitem}
\usepackage{placeins}
\newtheorem{Definition}{Definition}
\newcommand{\highlight}[1]{\noindent\textbf{#1}}
\definecolor{cvprblue}{rgb}{0.21,0.49,0.74}
\usepackage[pagebackref,breaklinks,colorlinks,allcolors=cvprblue]{hyperref}
\definecolor{iclrblue}{rgb}{0.21,0.49,0.74}
\definecolor{teaserred}{RGB}{180,10,56}

\definecolor{teaserblue}{RGB}{0,15,139}

\definecolor{uclablue}{RGB}{159, 195, 224}

\definecolor{uclagold}{RGB}{254,180,167}

\definecolor{grayred}{RGB}{232,237,205}

\definecolor{purpled}{RGB}{200, 162, 200}

\definecolor{myblue}{RGB}{108, 142, 191}

\definecolor{object_blue}{RGB}{103, 138, 207}
\newcommand{\objblue}[1]{\textcolor{object_blue}{\textbf{#1}}}
\definecolor{attribute_brown}{RGB}{211, 170, 108}
\newcommand{\attbrown}[1]{\textcolor{attribute_brown}{\textbf{#1}}}
\definecolor{relationship_green}{RGB}{56, 132, 36}
\newcommand{\relgreen}[1]{\textcolor{relationship_green}{\textbf{#1}}}
\definecolor{count_purple}{RGB}{196, 152, 176}
\newcommand{\coupurple}[1]{\textcolor{count_purple}{\textbf{#1}}}
\usepackage{color, colortbl}

\newcommand\best[1]{\textcolor{red}{\textbf{#1}}}

\newtcolorbox[auto counter, number within=section]{promptbox}[2][]{%
  colback=white, 
  colframe=myblue,  
  width=\textwidth,
  arc=2mm, 
  title={\normalsize\faInfoCircle\hspace{0.5em}#2},
  breakable,
  fonttitle=\bfseries\Large, 
  fontupper=\small, 
  drop shadow southeast, 
  top=2mm,
  bottom=2mm,
  before skip=3mm,
  after skip=3mm,
  boxrule=0.5mm,
  #1
}

\definecolor{iccvblue}{rgb}{0.21,0.49,0.74}
\definecolor{TealBlue}{rgb}{1.0, 0.97, 0.8}

\definecolor{verylightgray}{gray}{0.85}

\definecolor{mylightblue}{rgb}{0.2,0.45,0.85} 

\def\paperID{21052} 
\def\confName{CVPR}
\def\confYear{2026}

\title{Synthetic Curriculum Reinforces Compositional Text-to-Image Generation}

\author{
Shijian Wang\thanks{Equal contribution. Work done when Shijian internship at Xiaohongshu} \ $^{1,2,3}$
\quad Runhao Fu\footnotemark[1] \ $^{3,5}$
\quad Siyi Zhao$^{4}$
\quad Qingqin Zhan$^{6}$
\quad Xingjian Wang$^{6}$ \\
\quad Jiarui Jin\footnotemark[2] \ $^{2}$
\quad Yuan Lu\footnotemark[2] \ $^{2}$
\quad Hanqian Wu\thanks{Corresponding authors} \ $^{1}$
\quad Cunjian Chen$^{3}$ \\
\small $^{1}$Southeast University \quad $^{2}$Xiaohongshu Inc. \quad $^{3}$Monash University \\
\small \quad $^{4}$Shanghai Jiao Tong University \quad $^{5}$Anhui University \quad $^{6}$Independent Researcher
}

\begin{document}
\maketitle

\begin{abstract}
Text-to-Image (T2I) generation has long been an open problem, with compositional synthesis remaining particularly challenging. This task requires accurate rendering of complex scenes containing multiple objects that exhibit diverse attributes as well as intricate spatial and semantic relationships, demanding both precise object placement and coherent inter-object interactions. In this paper, we propose a novel compositional curriculum reinforcement learning framework named CompGen that addresses compositional weakness in existing T2I models. Specifically, we leverage scene graphs to establish a novel difficulty criterion for compositional ability and develop a corresponding adaptive Markov Chain Monte Carlo graph sampling algorithm. This difficulty-aware approach enables the synthesis of training curriculum data that progressively optimize T2I models through reinforcement learning. We integrate our curriculum learning approach into Group Relative Policy Optimization (GRPO) and investigate different curriculum scheduling strategies. Our experiments reveal that CompGen exhibits distinct scaling curves under different curriculum scheduling strategies, with easy-to-hard and Gaussian sampling strategies yielding superior scaling performance compared to random sampling. Extensive experiments demonstrate that CompGen significantly enhances compositional generation capabilities for both diffusion-based and auto-regressive T2I models, highlighting its effectiveness in improving the compositional T2I generation systems.
\end{abstract}

\section{Introduction}
\label{sec:intro}
Text-to-Image (T2I) generation has achieved remarkable progress in synthesizing visually compelling content from textual descriptions~\citep{ramesh2022hierarchical,saharia2022photorealistic,podell2023sdxl,balaji2022ediff}. 
\begin{figure}[htbp]
    \centering
    \vspace{-2mm}
    \includegraphics[width=\columnwidth,page=1]{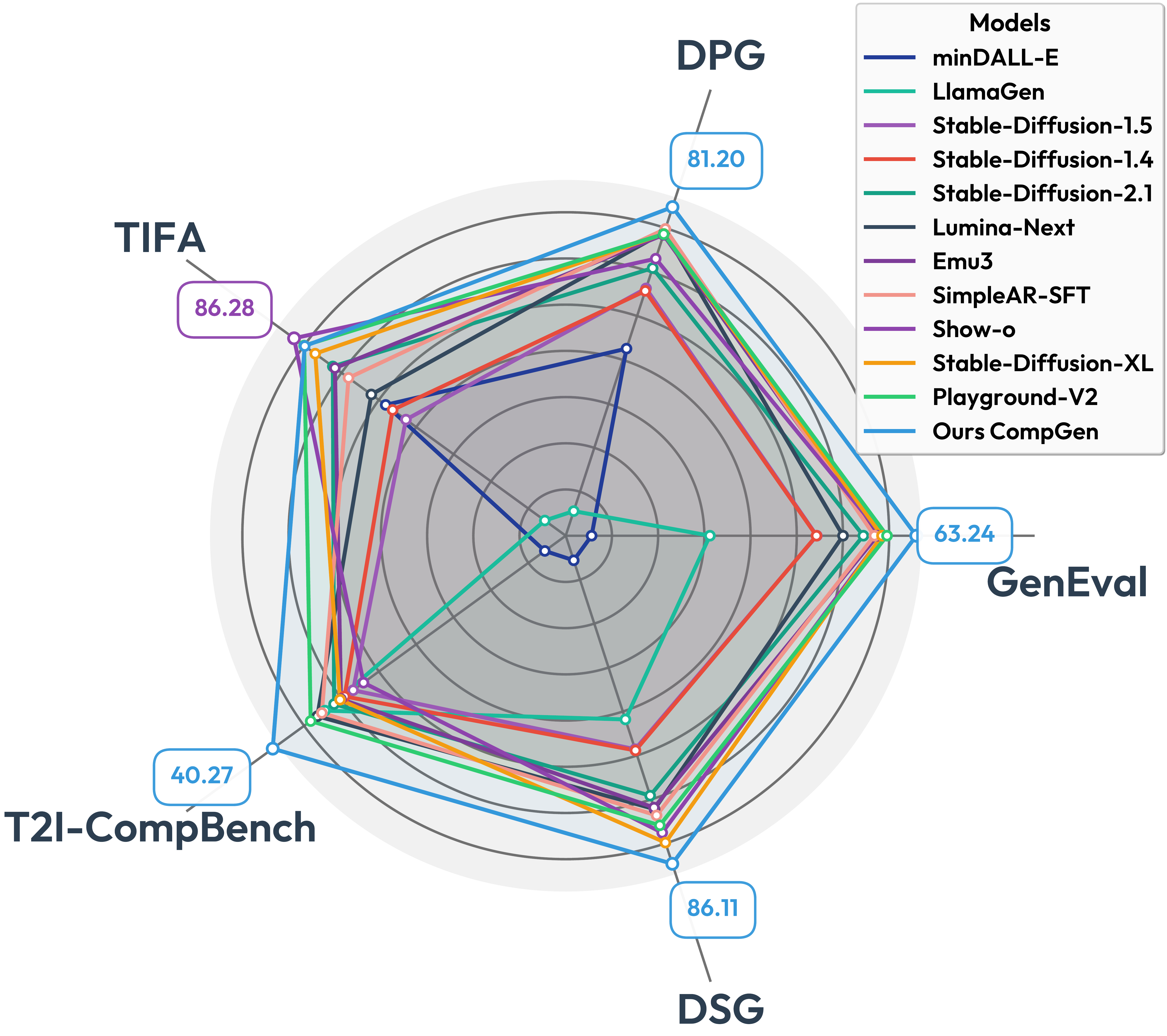}
    \vspace{-5mm}
    \caption{Overall performance of our CompGen, indicating that CompGen achieves state-of-the-art performance among models of the same scale.}
    \label{fig:radar}
    \vspace{-6mm}
\end{figure}
Despite these advances, current T2I models face significant limitations in compositional synthesis, particularly in accurately rendering complex scenes containing multiple objects with diverse attributes and intricate spatial relationships~\citep{liu2022compositional,nie2024compositional}.
Thus, compositional T2I generation with complex instructions \citep{huang2025t2i,hu2024ella} is still an open problem. 
To cope with this challenge, plenty of literature focuses on developing new network architectures such as attention models \citep{chefer2023attend,rassin2023linguistic,meral2024conform,kim2023dense}, or introducing intermediate structures like object layout \citep{chen2024training,dahary2024yourself,wang2024compositional}.

Unlike recent methods that require synthesized ground-truth images \citep{gao2024generate, sun2023dreamsync} or intermediate skeletons \citep{nie2024compositional} for supervised fine-tuning, our approach adopts a data-centric perspective that enhances compositional generalization through synthetic curriculum. Crucially, our method requires only textual prompts and applies reinforcement learning (RL) without the need for ground-truth image outputs.
However, large-scale RL training for compositional T2I generation faces significant stability challenges due to the heterogeneous nature of compositional capabilities required, which encompass object existence, attribute binding, relational understanding, and numerical counting.

\begin{figure*}[t]
    \centering
    \vspace{-6mm}
    \includegraphics[width=\textwidth]{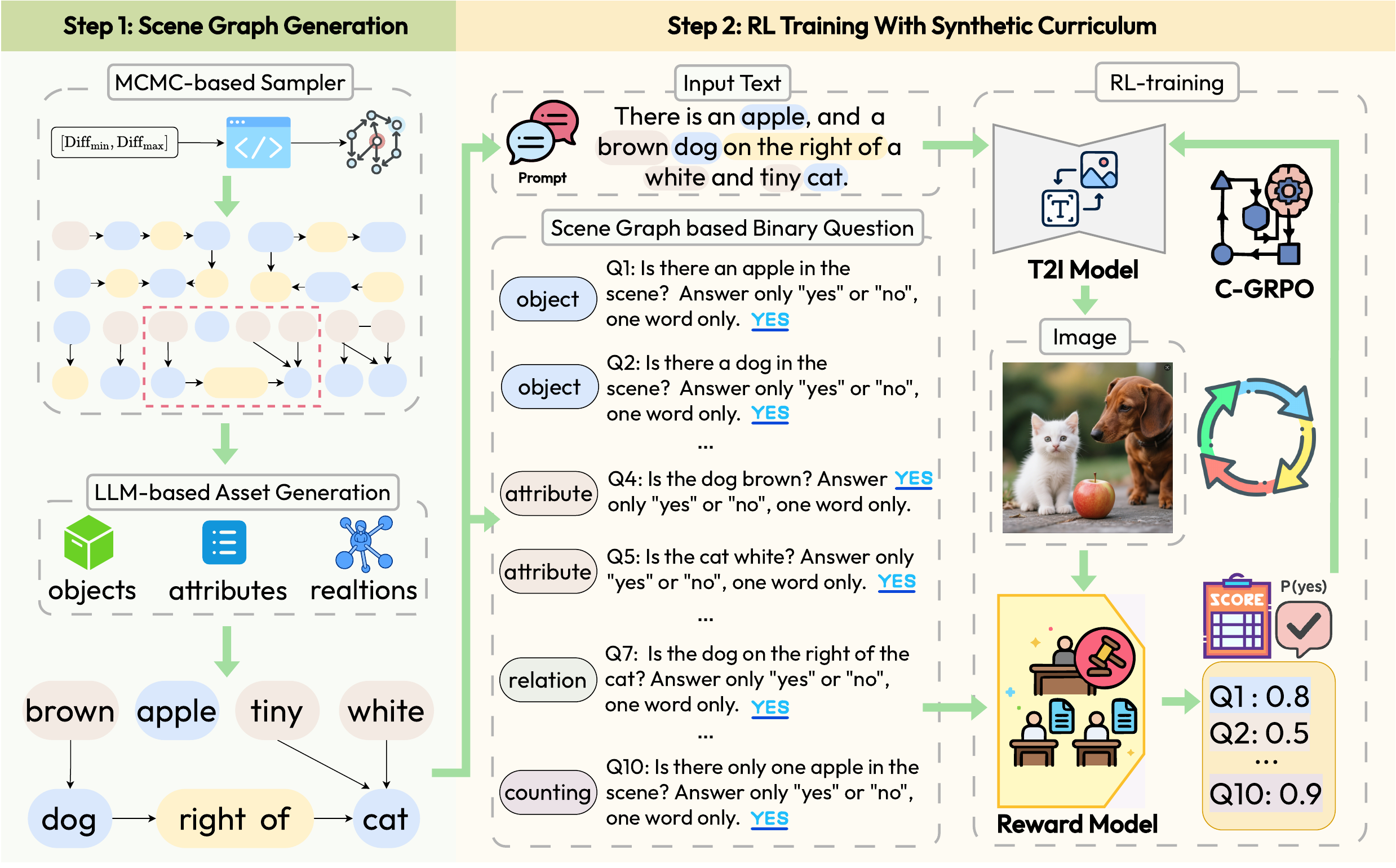}
    \vspace{-5mm}
    \caption{Overview of our CompGen framework, which is incentivized to construct a curriculum through end-to-end reinforcement learning without requiring ground-truth images.}
    \label{fig:overview}
    \vspace{-3mm}
\end{figure*}

To address this challenge, we propose \textbf{CompGen}, a novel compositional curriculum-based RL framework for T2I generation. 
CompGen draws inspiration from human cognitive development, which follows a curriculum learning progression: first mastering the recognition and generation of individual objects and their attributes within simple relational contexts, then gradually learning to understand and create complex multi-object compositions involving multiple relations.
Specifically, our approach leverages scene graphs \citep{krishna2017visual} as a compositional representation of visual scenes to systematically generate training data with controllable difficulty.
We introduce a novel difficulty criterion that quantifies compositional complexity based on scene graph structural properties, including entity count, attribute diversity, and relational interconnectedness. 
Leveraging this difficulty metric, we then develop an adaptive Markov Chain Monte Carlo (MCMC) sampling algorithm~\citep{geyer1992practical} to systematically generate scene graphs at specific difficulty levels, thereby enabling precise curriculum control throughout the training process.
For each sampled scene graph, we synthesize corresponding input text prompt for T2I generation and construct comprehensive visual question-answer pairs for assessments, namely object existence, object counting, attribute recognition, and relational understanding. 
These question-answer pairs subsequently serve as reward metrics within our RL framework, guiding the model toward improved compositional generation performance.

As presented in Figure~\ref{fig:radar}, our extensive experiments demonstrate that CompGen significantly strengthens compositional generation capabilities across both diffusion and auto-regressive T2I architectures. 
On five established compositional generation benchmarks — GenEval~\citep{ghosh2023geneval}, DPG~\citep{hu2024ella}, TIFA~\citep{hu2023tifa}, T2I-CompBench~\citep{huang2025t2i}, and DSG~\citep{cho2023davidsonian} — our approach consistently outperforms baseline models, achieving an average improvement of 11.72\% when applied to Stable-Diffusion-1.5 and 7.61\% when applied to SimpleAR. 
Moreover, our analysis reveals that curriculum scheduling critically impacts scaling behavior in curriculum-based GRPO training: easy-to-hard and Gaussian sampling strategies demonstrate superior performance and extended scaling potential compared to random sampling approaches.


\begin{figure}[t]
    \centering
    \includegraphics[width=0.7\columnwidth]{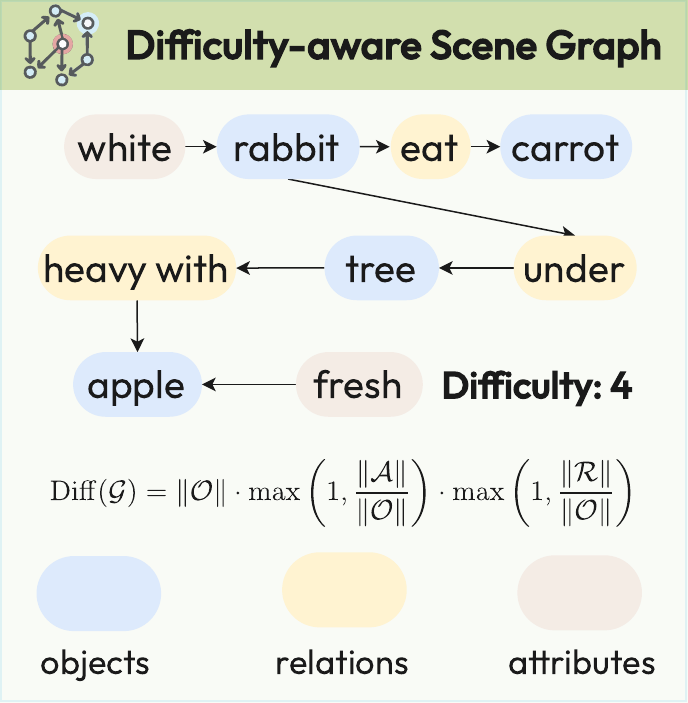}
    \caption{An illustrated example of scene graph corresponding to a specific difficulty level.}
    \label{fig:difficulty}
    \vspace{-5mm}
\end{figure}

\section{Related Work}
\label{sec:related}
Large text-to-image (T2I) generative models have attracted considerable attention in recent years and can be broadly categorized into two main families: diffusion-based models \citep{ramesh2022hierarchical,rombach2022high,podell2023sdxl,balaji2022ediff} and auto-regressive models \citep{ramesh2021zero,yu2022scaling,chang2023muse}.
Beyond purely improving visual quality, recent investigations \citep{wang2024phased,wang2024your} have focused on enhancing prompt-following capabilities, particularly for compositional prompts. 
However, existing T2I models often struggle with compositional understanding, leading to issues such as object omission and incorrect attribute binding~\citep{okawa2023compositional,huang2025t2i}.
To address these limitations, recent efforts to improve compositional alignment can be grouped into three main categories.
\textbf{Attention-based methods} modify attention maps within the UNet architecture to enforce object presence and spatial separation~\citep{chefer2023attend,rassin2023linguistic,feng2022training,meral2024conform,kim2023dense}. 
For instance, DenseDiffusion~\citep{kim2023dense} adjusts attention scores in both cross-attention and self-attention layers to ensure object features align with specified image regions, while CONFORM~\citep{meral2024conform} strengthens associations between relevant objects and attributes through contrastive objectives. 
However, these approaches face scalability and computational efficiency limitations as they operate only during inference.
\textbf{Planning-based methods} utilize intermediate structures such as object layouts — either manually defined~\citep{chen2024training,dahary2024yourself,wang2024compositional} or generated by large language models~\citep{gani2023llm,lian2023llm} — to guide image synthesis. 
Some approaches incorporate additional modules like visual question-answering models or captioning models for refinement~\citep{yang2024mastering,wu2024self,wen2023improving}; however, these additions increase inference costs and may still suffer from incorrect attribute bindings due to inherent model limitations.
\textbf{Learning-based methods} focus on training-time improvements, including fine-tuning diffusion models with vision-language supervision~\citep{wang2024compositional,wang2022semantic,fan2023dpok} or employing reinforcement learning techniques~\citep{fan2023dpok,black2023training}. 
Caption-guided optimization represents another promising direction in this category~\citep{fang2023boosting,ma2024exploring}.

Unlike previous approaches that require additional inputs or architectural modifications, our method adopts a data-centric strategy to enhance compositional generalization through synthetic curriculum-based RL, without increasing inference costs or altering model architecture.

\section{Scene Graph as a Difficulty Measurer}
\label{sec:pre}
The core principle of CompGen is to progressively develop compositional generation capabilities through curriculum learning, advancing from simple to complex samples. The central challenge in this approach lies in establishing a principled framework for defining and quantifying the difficulty of compositional samples.
Following the formalized image representation framework introduced by \citet{krishna2017visual}, scene graphs (composed of three types of nodes: objects, attributes, relationships, and directed edges) provide a structured approach to capturing compositional complexity. We therefore propose to measure sample difficulty through the compositional complexity inherent in scene graph structures.
Formally, we provide the definition of difficulty based on scene graphs as follows:
\begin{Definition}
[Scene Graph Formulated Difficulty] Given a scene graph $\mathcal{G}=(\mathcal{O}, \mathcal{A}, \mathcal{R})$ where $\mathcal{O}$ is the set of objects, $\mathcal{A}$ is the set of attributes associated with the objects, and $\mathcal{R}$ is the set of relations, we measure the difficulty of $\mathcal{G}$ as:
\begin{equation}
\label{eqn:difficulty}
\emph{\text{Diff}}(\mathcal{G}) = \|\mathcal{O}\| \cdot \max \Bigg(1, \frac{\|\mathcal{A}\|}{\|\mathcal{O}\|}\Bigg) \cdot \max \Bigg(1, \frac{\|\mathcal{R}\|}{\|\mathcal{O}\|}\Bigg),
\end{equation}
where, the total number of objects is $\|\mathcal{O}\|$, the average attribute density is $\|\mathcal{A}\| / \|\mathcal{O}\|$, and the average relational connectivity is $\|\mathcal{R}\| / \|\mathcal{O}\|$ per object. 
\label{def:graph}
\end{Definition}
Eq.~(\ref{eqn:difficulty}) demonstrates that the computational difficulty of $\mathcal{G}$ is determined by three key factors identified by the total number of objects, the average attribute density, and the average relational connectivity.
Consequently, the overall difficulty exhibits a positive correlation with the intrinsic complexity of the scene graph, reflecting both its structural density and semantic richness.
We provide an illustrated case of our scene graph of a specific difficulty level in Figure~\ref{fig:difficulty}. 
We also empirically compare the CompGen performance of using Eq.~(\ref{eqn:difficulty}) with other formulations in Section~\ref{sec:abla_difficulty}.

\section{RL Training with Synthetic Curriculum}
Given the difficulty measure introduced in Definition~\ref{def:graph}, we can construct a mapping function $\{\mathcal{G}\}\rightarrow \mathbb{R}^+$ over the space of scene graphs $\mathcal{G}$.
Building upon this foundation, we formalize our problem as follows:
\begin{Definition}[Synthetic Curriculum-based RL]
\label{def:t2i_difficulty}
Given a target difficulty range $[\text{\emph{Diff}}_\text{\emph{min}}, \text{\emph{Diff}}_\text{\emph{max}}]$ where $0 < \text{\emph{Diff}}_\text{\emph{min}} \leq \text{\emph{Diff}}_\text{\emph{max}}$, 
a synthetic curriculum for T2I can be formulated as a T2I task with an input text $T$ and a reward $r$ corresponding to the output image $I$.
Therefore, our methodology follows a structured pipeline: we first construct a scene graph $\mathcal{G}$ satisfying the difficulty constraint $\text{\emph{Diff}}_\text{\emph{min}} \leq \text{\emph{Diff}}(\mathcal{G}) \leq \text{\emph{Diff}}_\text{\emph{max}}$, then derive the corresponding input text $T$ from this graph. 
These text prompts are processed by a T2I model to generate images $I$s, which are subsequently evaluated by the reward function $R$. 
The resulting rewards enable optimization of the T2I model within a RL framework, ensuring adherence to the specified difficulty constraints.
\end{Definition}

Based on the aforementioned problem formulation, our CompGen framework operates through a two-stage methodology, as demonstrated in Figure~\ref{fig:overview}. 
Specifically, the framework first generates a scene graph with a specific range of difficulty levels to construct a synthetic curriculum, and subsequently trains T2I models through curriculum-based RL.

\subsection{Scene Graph Generation via Adaptive Markov Chain Monte Carlo Sampling}
Given the target range of difficulty levels $\text{Diff}_\text{min}$ and $\text{Diff}_\text{max}$, we define the constrained generation problem as the task of sampling scene graphs $\mathcal{G}$ such that $\text{Diff}_\text{min} \leq \text{Diff}(\mathcal{G}) \leq \text{Diff}_\text{max}$.
A brute-force enumeration of all possible scene graphs to identify those satisfying the difficulty constraints is computationally intractable due to the exponentially large and high-dimensional nature of the graph space. 
To efficiently navigate this combinatorial landscape, we reformulate the task as an iterative sampling problem and employ a targeted sampling strategy based on Markov Chain Monte Carlo (MCMC) \citep{brooks1998markov}. 

Our approach begins with an initial scene graph $\mathcal{G}_0$ and iteratively refines it through systematic modifications to discover graph structures that satisfy the specific difficulty constraints. 
In practice, we initialize $\mathcal{G}_0$ as a minimally complex baseline graph, typically comprising a randomly selected small number of object nodes, thereby establishing a neutral starting point for the MCMC sampling process.
To enable this graph sampling process, we define a set of two kinds of reversible graph transformation operations, namely $\mathcal{T}_\text{add}$ and $\mathcal{T}_{\text{delete}}$. The addition operation, $\mathcal{T}_\text{add}$, introduces a new element (i.e., an attribute/relation node and its associated edges), while its inverse, the delete operation $\mathcal{T}_{\text{delete}}$, eliminates an existing element (i.e., a node and all its incident edges). 
At each MCMC sampling step, a candidate graph $\mathcal{G}'$ is proposed from the current state $\mathcal{G}$ by randomly selecting and applying one of these transformations, thereby defining our proposal distribution $q(\mathcal{G}'|\mathcal{G})$. 
Crucially, these operations are designed to maintain detailed balance through symmetric transition probabilities; specifically, the probability of proposing to add a particular element to graph $\mathcal{G}$ equals the probability of proposing to delete that identical element from the resulting graph $\mathcal{G}'$.

To systematically guide the MCMC sampling process toward the target difficulty range, we define an energy function $\text{Energy}(\mathcal{G})$ that quantifies the extent to which a graph's difficulty deviates from the specified constraints:
\begin{equation}
\label{eq:energy}
\begin{aligned}
\text{Energy}(\mathcal{G}) &:= \text{Dist}(\text{Diff}(\mathcal{G}), [\text{Diff}_\text{min},\text{Diff}_\text{max}]) \\
&= \begin{cases}
    \text{Diff}_\text{min} - \text{Diff}(\mathcal{G}), & \text{if } \text{Diff}(\mathcal{G}) < \text{Diff}_\text{min} \\
    \text{Diff}(\mathcal{G}) - \text{Diff}_\text{max}, & \text{if } \text{Diff}(\mathcal{G}) > \text{Diff}_\text{max} \\
    0, & \text{otherwise}
\end{cases}
\end{aligned}
\end{equation}
The energy function evaluates to zero when a graph's difficulty falls within the target range and assumes positive values otherwise, thereby establishing the optimization objective of identifying graphs with minimal (ideally zero) energy. 
We subsequently employ the Metropolis-Hastings algorithm \citep{chib1995understanding} to make stochastic acceptance decisions for proposed transformations. 
Transitions to lower-energy states (i.e., configurations closer to the target difficulty range) are unconditionally accepted, while transitions to higher-energy states are accepted with probability determined by the Metropolis criterion to facilitate exploration and prevent convergence to suboptimal local minima.
The acceptance probability is formally defined as:
\begin{equation}
\label{eqn:acceptance}
\text{Acc}(\mathcal{G}'|\mathcal{G}) = \min\left(1, \frac{\pi(\mathcal{G}')q(\mathcal{G}|\mathcal{G}')}{\pi(\mathcal{G})q(\mathcal{G}'|\mathcal{G})}\right),
\end{equation}
where $\pi(\mathcal{G}) \propto \exp(-\text{Energy}(\mathcal{G})/\tau)$ is the target distribution that favors low-energy graphs. Due to the symmetric design of our proposal distribution $q(\mathcal{G}'|\mathcal{G})$, the proposal ratio $\frac{q(\mathcal{G}|\mathcal{G}')}{q(\mathcal{G}'|\mathcal{G})}$ is unity, simplifying the acceptance probability to depend only on the change in energy $\text{Acc}(\mathcal{G}'|\mathcal{G}) = \min\left(1, \exp\left(\frac{\text{Energy}(\mathcal{G}) - \text{Energy}(\mathcal{G}')}{\tau}\right)\right)$.
Here, temperature parameter $\tau$ governs the trade-off between exploration and exploitation, thereby rendering our sampling algorithm adaptive to the current graph state and energy landscape.

Concretely, we employ an adaptive temperature schema analogous to simulated annealing.
We initialize with a high temperature $\tau$ to facilitate extensive exploration of the graph space and progressively reduce it throughout the sampling process. 
This annealing schedule enables the sampler to initially identify promising regions of the solution space before progressively refining the search to achieve precise convergence on graphs that satisfy the specified difficulty constraints. 
Our empirical evaluation demonstrates that this approach outperforms baseline methods in both sampling efficiency and graph diversity, as detailed in Appendix~\ref{app:mcmc_exp}. 
We additionally investigate the effects of various initialization strategies on sampling performance and node type diversity in Appendix~\ref{app:g0}. 
The complete algorithmic procedure is presented in Algorithm~\ref{alg:constrained_sampler} in Appendix~\ref{app:algo}.

\subsection{Curriculum-based Group Relative Policy Optimization}
\label{sec:CGRPO}
Having obtained a scene graph $\mathcal{G}$ through the constrained sampling process described above, we instantiate the abstract graph structure by randomly sampling concrete semantic values from our comprehensive library of scene graph assets.
This library comprises diverse objects, attributes, and relationships constructed using an LLM with implementation details provided in Appendix~\ref{app:asset} and specific prompts detailed in Appendix~\ref{app:asset_prompt}.

Upon instantiation of the scene graph, we generate the corresponding input text $T$ for the T2I model and formulate an appropriate reward function for the subsequent training phase (as exemplified in Figure~\ref{fig:overview}).
The reward function $R$ takes as input the binary question-answer pairs and the generated images, then outputs a scalar reward value $r\in\mathcal{R}$ to guide the compositional learning process.

\highlight{Generation of Input Text $T$.}
CompGen transforms structured scene graphs into natural language prompts while preserving their original compositional difficulty. 
Given a scene graph $\mathcal{G} = (\mathcal{O}, \mathcal{A}, \mathcal{R})$ with associated difficulty $\text{Diff}(\mathcal{D}) \in [\text{Diff}_\text{min}, \text{Diff}_\text{max}]$, we employ constrained LLM based generation with strict constraints to produce descriptions that exactly match the graph's specifications. 
The generation process integrates the linguistic capabilities of LLMs with systematic validation mechanisms to maintain strict fidelity to the source graph. 
This is achieved through three core components: (i) mandatory inclusion constraints that ensure exact matching of all objects and attributes, (ii) structural validation that preserves relational dependencies, and (iii) multi-stage content verification that maintains semantic integrity. 
This approach guarantees that the generated text prompt comprehensively reflects all elements of the original scene graph while preserving its compositional difficulty. 
In our implementation, we utilize Deepseek-V3 model \citep{liu2024deepseek} as the underlying LLM, with detailed prompts provided in Appendix~\ref{app:input_text_prompt}.

\begin{table*}[t]
\centering
\caption{Comparison of different T2I models on compositional generation benchmarks. Models with \colorbox{verylightgray}{gray} background are the baseline models that our method builds upon, while those with \colorbox{TealBlue}{yellow} background are our trained models. The best performance among all models are marked in \textcolor{red}{\textbf{red}} and the performance improvement of our trained models over the baseline models are marked in \textcolor{ForestGreen}{\textbf{↑green}}.}
\captionsetup{skip=4pt}
\renewcommand{\arraystretch}{1.5}
\fontsize{8pt}{10pt}\selectfont
\resizebox{\textwidth}{!}{
\begin{tabular}{lccccccc}
\toprule
\specialrule{0.5pt}{0pt}{0pt}
\rowcolor{lightgray}
\textbf{Model} & \textbf{\# Params} & \textbf{GenEval} & \textbf{DPG} & \textbf{TIFA} & \textbf{T2I-CompBench} & \textbf{DSG} & \textbf{Avg.} \\
\specialrule{0.5pt}{0pt}{0pt}
\midrule
\multicolumn{8}{c}
{\cellcolor{grayred}\textbf{\textit{Diffusion}}} \\
Stable-Diffusion-1.4 & 0.9B & 42.04\% & 61.89\% & 79.14\% & 30.80\% & 61.71\% & 55.12\% \\
\rowcolor{verylightgray}
Stable-Diffusion-1.5 & 0.9B & 42.08\% & 62.24\% & 78.67\% & 29.94\% & 61.57\% & 54.90\% \\
Stable-Diffusion-2.1 & 0.9B & 50.00\% & 65.47\% & 82.00\% & 32.01\% & 68.09\% & 59.51\% \\
Playground-V2 & 2.6B & 59.00\% & 74.54\% & 86.20\% & 36.13\% & 74.54\% & 66.08\% \\
Stable-Diffusion-XL & 2.6B & 55.87\% & 74.65\% & 83.50\% & 31.30\% & 83.40\% & 65.74\% \\

Lumina-Next & 1.7B & 46.00\% & 75.66\% & 79.98\% & 34.57\% & 70.61\% & 61.36\% \\
\midrule
\multicolumn{8}{c}
{\cellcolor{uclablue}\textbf{\textit{AutoRegressive}}} \\
LlamaGen & 0.8B & 31.28\% & 42.92\% & 75.03\% & 33.26\% & 58.30\% & 48.16\% \\
Show-o & 1.3B & 56.00\% & 67.27\% &  \textcolor{red}{\textbf{86.28\%}} & 29.00\% & 77.00\% & 63.11\% \\
\rowcolor{verylightgray}
SimpleAR-SFT & 0.5B & 53.00\% & 78.48\% & 81.06\% & 33.76\% & 71.98\% & 63.66\% \\
Emu3 & 14B & 54.00\% & 74.19\% & 81.86\% & 31.20\% & 70.31\% & 62.31\% \\
minDALL-E   & 1.3B & 23.00\% & 55.23\% & 79.40\% & 18.98\% & 45.63\% & 44.45\% \\
\midrule
\multicolumn{8}{c}{\cellcolor{purpled}\textbf{\textit{Ours}}} \\
\rowcolor{TealBlue}
Stable-Diffusion-1.5 w/ ours 
& 0.9B 
& 53.88\% \textcolor{ForestGreen}{\textbf{(↑11.80\%)}} 
& 78.67\% \textcolor{ForestGreen}{\textbf{(↑16.43\%)}} 
& 85.71\% \textcolor{ForestGreen}{\textbf{(↑7.04\%)}} 
& 37.68\% \textcolor{ForestGreen}{\textbf{(↑7.74\%)}} 
& 77.16\% \textcolor{ForestGreen}{\textbf{(↑15.59\%)}}  
& 66.62\% \textcolor{ForestGreen}{\textbf{(↑11.72\%)}} \\
\rowcolor{TealBlue}
SimpleAR w/ ours 
& 0.5B 
& \textcolor{red}{\textbf{63.24\%}} \textcolor{ForestGreen}{\textbf{(\textbf{↑10.24\%})}} 
& \textcolor{red}{\textbf{81.20\%}} \textcolor{ForestGreen}{\textbf{(\textbf{↑2.72\%})}} 
& 85.53\% \textcolor{ForestGreen}{\textbf{(\textbf{↑4.47\%})}} 
& \textcolor{red}{\textbf{40.27\%}} \textcolor{ForestGreen}{\textbf{(\textbf{↑6.51\%})}} 
& \textcolor{red}{\textbf{86.11\%}} \textcolor{ForestGreen}{\textbf{(\textbf{↑14.13\%})}}  
& \textcolor{red}{\textbf{71.27\%}} \textcolor{ForestGreen}{\textbf{(\textbf{↑7.61\%})}} \\
\bottomrule
\end{tabular}}
\label{tab:gen_models}
\end{table*}

\highlight{Generation of Curriculum.}
Given input text prompts $T$, any T2I model generates corresponding images $I$s.
We evaluate the compositional accuracy of these generated images through structured question-answer pairs that are systematically constructed from the sampled scene graphs. 
Specifically, we adopt a programmatic approach \citep{gao2024generate} to generate precise and comprehensive binary questions that exhaustively cover all structural elements of the scene graphs.
Since each scene graph $\mathcal{G} = (\mathcal{O}, \mathcal{A}, \mathcal{R})$ encodes objects, attributes, and relations, with object cardinality representing an additional critical dimension, we design four complementary question categories to achieve comprehensive coverage:
(i) object verification questions ($Q_{\text{object}}$) to assess the presence of each object $o \in \mathcal{O}$;
(ii) count verification questions ($Q_{\text{count}}$) to verify the correct number of occurrences for repeated objects in $\mathcal{O}$;
(iii) attribute validation questions ($Q_{\text{attribute}}$) to confirm the presence of each attribute $a \in \mathcal{A}$ for its corresponding objects;
(iv) relation confirmation questions ($Q_{\text{relation}}$) to verify the existence of each relation $r \in \mathcal{R}$ between specific object pairs.

\highlight{Generation of Reward $r\in\mathcal{R}$.}
Building upon the binary question-answer pairs constructed above and inspired by VQAScore~\citep{lin2024evaluating}, we develop an automated reward mechanism for T2I model training. 
Specifically, we leverage a multimodal LLM (MLLM) to evaluate image-text alignment by computing the predicted probability of answering ``yes'' to each binary question as a fine-grained reward score.
We take the average of the VQA scores for all questions corresponding to each image as the reward signal for reinforcement training, enabling the T2I model to perform policy updates.
For ease of notation, we use $p_\text{reward}(\cdot)$ to denote the MLLM in use.
In our implementation, we adopt LLaVA-v1.6-13B~\citep{liu2023visual} as the reward model due to its robust performance on visual question answering tasks. 
To validate the generalizability of our approach, we conduct comprehensive ablation studies with alternative MLLMs in Section~\ref{sec:abla_reward}.

\highlight{Model Training.}
During training, we integrate Curriculum Learning with Group Relative Policy Optimization (GRPO) to progressively enhance the T2I model's compositional generation ability. 
The core idea of Curriculum-based GRPO (C-GRPO) is to dynamically adjust the emphasis on different compositional difficulty levels throughout training, enabling the model to first master simpler concepts before tackling more complex compositions.
For each generated image $I^{(i)}$, we evaluate its compositional quality using binary question-answer pairs through our reward model as $r_j^{(i)} = p_{\text{reward}}(\text{answer}_j | I^{(i)}, \text{question}_j)$,
where $r_j^{(i)} \in [0, 1]$ represents the reward model's confidence that image $I^{(i)}$ correctly answers the $j$-th binary question corresponding to a specific compositional difficulty level.

Following \citet{parashar2025curriculum}, we employ curriculum scheduling strategies to weight these rewards based on training progress.
We explore two scheduling approaches (as detailed in Appendix~\ref{app:strategy}), namely Easy-to-Hard scheduling and Gaussian scheduling. 
The curriculum-weighted reward at training step $t$ is computed as
$\widehat{r}_j^{(i)}(t)=\sum^{||\text{Diff}||}_{j'=1} \widehat{p}(t,j')\cdot r_j^{(i)}$, where $||\text{Diff}||$ denotes the number of compositional difficulty levels, and $\widehat{p}(t,j')$ represents the curriculum sampling probability for difficulty level $j'$ at step $t$.
We derive the detailed formulation for $\widehat{p}(t,j')$ in Appendix~\ref{app:strategy}.
The overall reward for image $I^{(i)}$ is computed as the average score across all the sampled questions:
$\widehat{r}^{(i)}(t) = \frac{1}{M} \sum_{j=1}^M \widehat{r}_j^{(i)}(t)$, where $M$ is the number of samples.
This reward design ensures that at each training stage, the model focuses on the appropriate difficulty level while gradually progressing toward more complex compositions.

The curriculum-aware advantages at step $t$ are normalized within each group of $G$ images as:
\begin{equation}
\label{eqn:advantage}
A_i(t) = \frac{\widehat{r}^{(i)}(t) - \text{Mean}(\{\widehat{r}^{(k)}(t)\}_{k=1}^G)}{\text{Std}(\{\widehat{r}^{(k)}(t)\}_{k=1}^G)},
\end{equation}
where $\text{Mean}(\{\widehat{r}^{(k)}(t)\}_{k=1}^G)$ and $\text{Std}(\{\widehat{r}^{(k)}(t)\}_{k=1}^G)$ denote the mean and standard deviation of rewards within the group.

For each sampled input text prompt $T$, C-GRPO generates $G$ distinct images $\{I^{(1)}, I^{(2)}, \ldots, I^{(G)}\}$ using the current policy $p_{\theta_{\text{old}}}$. 
The policy at step $t$ is optimized by maximizing:
\begin{equation}
\label{eqn:loss}
\begin{aligned}
&\mathcal{J}_{\text{C-GRPO}}(\theta) = \mathbb{E}_{T} \Bigg[
\frac{1}{G} \sum_{i=1}^G \Big( \min \Big( \frac{\pi_\theta}{\pi_{\theta_{\text{old}}}} A_i(t), \\
&\text{clip}\Big( \frac{\pi_\theta}{\pi_{\theta_{\text{old}}}} , 1 - \epsilon, 1 + \epsilon \Big) A_i(t) \Big) - \beta \, \text{KL}\Big(p_\theta(\cdot | T) \Big\| p_{\text{ref}}(\cdot | T)\Big) \Big) \Bigg]
\end{aligned}
\end{equation}
where $\pi_\theta = p_\theta(I^{(i)} | T)$ denotes the probability of generating image $I^{(i)}$ given text prompt $T$ under the current policy, $\pi_{\theta_{\text{old}}} = p_{\theta_{\text{old}}}(I^{(i)} | T)$ is the probability under the old policy, $\epsilon$ and $\beta$ are hyperparameters for clipping and KL regularization respectively, and $p_{\text{ref}}$ is the reference policy.

\section{Experiment}
\label{sec:exp}

\subsection{Experimental Setup}
\label{sec:setup}
\highlight{Benchmark Datasets and Metrics.}
To thoroughly assess the compositional generation capability of models trained with our CompGen framework, we evaluate on the following five compositional T2I benchmarks:
(i) Geneval~\citep{ghosh2023geneval}, 
(ii) T2I-CompBench~\citep{huang2025t2i}, 
(iii) TIFA~\citep{hu2023tifa}, 
(iii) DPG-Bench~\citep{hu2024ella},
(iv) DSG~\citep{cho2023davidsonian}.
Specifically, for T2I-CompBench, we report the performance on its complex compositions task, while for the remaining benchmarks, we report the average performance across their respective subtasks. Further details on the datasets and metrics are available in Appendix~\ref{app:dataset}.

\begin{figure*}[htbp!]
    \begin{center}
    \vspace{-6mm}
    \centerline{\includegraphics[width=\textwidth]{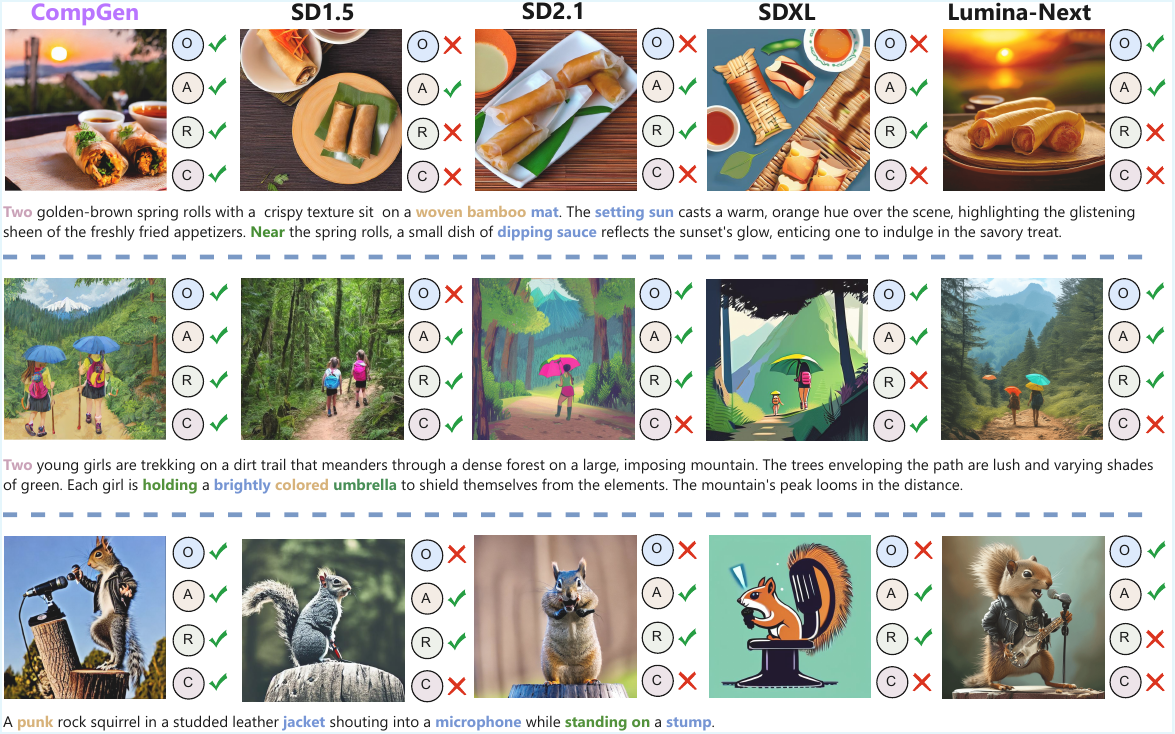}}
    \caption{Qualitative comparison of our CompGen with other strong text-to-image generation models (SD1.5, SD2.1, SDXL, and Lumina-Next). Within each prompt, we color the elements for which at least one model makes an error: the object in \objblue{blue}, the attribute in \attbrown{brown}, the relationship in \relgreen{green}, and the count in \coupurple{purple}.
    \raisebox{-0.4ex}{\includegraphics[height=1em]{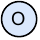}}, 
    \raisebox{-0.4ex}{\includegraphics[height=1em]{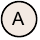}}, 
    \raisebox{-0.4ex}{\includegraphics[height=1em]{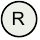}}, 
    \raisebox{-0.4ex}{\includegraphics[height=1em]{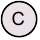}} denote Object, Attribute, Relationship, and Count, respectively. A
    \raisebox{-0.4ex}{\includegraphics[height=1em]{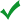}} indicates correct generation, while a   
    \raisebox{-0.4ex}{\includegraphics[height=1em]{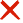}} indicates an error. Additional examples appear in Appendix~\ref{app:vis}.}
    \label{fig:compare}
    \end{center}
    \vspace{-3mm}
\end{figure*}

\highlight{Baseline Models.}
To validate the effectiveness of our proposed CompGen method, we conduct comprehensive comparisons against several state-of-the-art T2I generation models. These include diffusion-based T2I models such as Stable Diffusion 1.4/1.5/2.1~\citep{rombach2022high}, Playground v2~\citep{playground-v2}, Stable Diffusion XL~\citep{podell2023sdxl} and LUMINA-NEXT~\citep{zhuo2024lumina}; as well as auto-regressive T2I models such as Show-o~\citep{xie2024show}, Emu3~\citep{sun2023emu}, SimpeAR~\citep{wang2025simplear}, LLamaGen~\citep{sun2024autoregressive} and minDALL-E~\citep{kim2021mindall}.
We provide detailed descriptions for each baseline in  Appendix~\ref{app:models}.

\highlight{Implementation Details of CompGen.}
To generate high-quality training data for compositional RL, we employ our CompGen framework to construct 10K samples, which are evenly distributed across difficulty levels ranging from 1 to 10 according to our difficulty metric in Definition~\ref{def:graph}. We provide more detailed explorations of the difficulty measure in Section~\ref{sec:abla_difficulty} and data difficulty distribution in Appendix~\ref{app:ablation}. 
To demonstrate the effectiveness and generalizability of CompGen, we apply it to two prominent T2I architectures representing different generation paradigms: the diffusion-based Stable-Diffusion-1.5~\citep{rombach2022high} and the auto-regressive SimpleAR-SFT~\citep{sun2024autoregressive}.
We provide the training details of CompGen in Appendix~\ref{app:setup}.


\subsection{Performance Comparisons and Analysis}
\label{sec:performance}

\begin{table*}[tbp]
\vspace{-2mm}
\caption{Investigation on reward models using Stable-Diffusion-1.5 as backbone. Our adopted model is highlighted in \colorbox{TealBlue}{yellow}. The best performance is marked in \textcolor{red}{\textbf{red}}.}
\vspace{-2mm}
\centering
\fontsize{8pt}{10pt}\selectfont
\resizebox{0.9\textwidth}{!}{
\renewcommand{\arraystretch}{1.3}
\begin{tabular}{l ccccccc}
\toprule
\specialrule{0.5pt}{0pt}{0pt}
\rowcolor{lightgray}
\textbf{Model} & \textbf{Reward Model} & \textbf{GenEval} & \textbf{DPG} & \textbf{TIFA} & \textbf{T2I-CompBench} & \textbf{DSG} & \textbf{Avg.} \\
\specialrule{0.5pt}{0pt}{0pt}
\midrule
Stable-Diffusion-1.5 & -- & 42.08\% & 62.24\% & 78.67\% & 29.94\% & 61.57\% & 54.90\% \\
Stable-Diffusion-1.5 w/ VQAScore~\citep{lin2024evaluating} & LLaVA-v1.6-13B & 44.02\% & 73.41\% & 80.19\% & 36.36\% & 73.23\% & 61.44\% \\
\hline
\multirow{4}{*}{CompGen (Stable-Diffusion-1.5)} & CLIP-FlanT5-XXL & 45.11\% & 71.26\% & 81.42\% & 31.60\% & 73.74\% & 60.63\% \\
\cline{2-2}
& InstructBLIP & 42.04\% & 64.00\% & 76.29\% & \best{39.21\%} & 64.58\% & 57.22\% \\
\cline{2-2}
& LLaVA-v1.5-13B & 49.23\% & 74.54\% & 84.84\% & 37.43\% & 75.97\% & 64.40\% \\
\cline{2-2}
& \cellcolor{TealBlue} LLaVA-v1.6-13B & \cellcolor{TealBlue} \best{53.88\%} & \cellcolor{TealBlue} \best{78.67\%} & \cellcolor{TealBlue} \best{85.71\%} & \cellcolor{TealBlue} 37.68\% & \cellcolor{TealBlue} \best{77.16\%} & \cellcolor{TealBlue} \best{66.62\%} \\
\bottomrule
\end{tabular}}
\label{tab:reward_impact}
\end{table*}

\begin{table}[tbp]
\caption{
Investigation on different \textit{difficulty measures} using Stable-Diffusion-1.5 with 10K training data.
Our proposed difficulty measure is marked with \colorbox{TealBlue}{yellow} background.
The best performance for each benchmark is marked in \textcolor{red}{\textbf{red}}.
$\|\mathcal{O}\|$ is the number of objects in the scene graph, $\|\mathcal{A}\|$ is the number of attributes and
$\|\mathcal{R}\|$ is the number of relations.
}
\vspace{-2mm}
\centering
\resizebox{\columnwidth}{!}{
\fontsize{7pt}{9pt}\selectfont
\renewcommand{\arraystretch}{1.5}
\begin{tabular}{l cccccc}
\toprule
\specialrule{0.5pt}{0pt}{0pt}
\rowcolor{lightgray}
\textbf{Difficulty Measure} & \textbf{GenEval} & \textbf{DPG} & \textbf{TIFA} & \textbf{T2I-CompBench} & \textbf{DSG} \\
\specialrule{0.5pt}{0pt}{0pt}
\midrule
$(|\|\mathcal{O}\| + \|\mathcal{A}\| + \|\mathcal{R}\|)$~\citep{gao2024generate} 
& 50.12\% 
& 72.59\% 
& 79.40\% 
& 37.00\% 
& 71.21\% \\
${(\|\mathcal{O}\| + \|\mathcal{R}\|)}/2$~\citep{chen2024makes} 
& 48.83\% 
& 72.91\% 
& 75.24\% 
& 35.59\% 
& 74.33\% \\
\cellcolor{TealBlue} Ours 
& \cellcolor{TealBlue} \textcolor{red}{\textbf{53.88\%}} 
& \cellcolor{TealBlue} \textcolor{red}{\textbf{78.67\%}} 
& \cellcolor{TealBlue} \textcolor{red}{\textbf{85.71\%}} 
& \cellcolor{TealBlue} \textcolor{red}{\textbf{37.68\%}} 
& \cellcolor{TealBlue} \textcolor{red}{\textbf{77.16\%}} \\
\bottomrule
\end{tabular}}
\label{tab:difficulty_measure}
\end{table}

\highlight{CompGen significantly improves T2I compositional generation capabilities across both diffusion and auto-regressive architectures.}
As shown in Table~\ref{tab:gen_models}, CompGen delivers substantial improvements on compositional generation benchmarks. Applied to Stable-Diffusion-1.5 (0.9B parameters), it achieves +11.72 percentage points average improvement, surpassing both the stronger Stable-Diffusion-2.1 (59.51\%) and the larger Playground-V2 (66.08\%, 2.6B parameters).
On auto-regressive models, CompGen demonstrates similar versatility. Applied to SimpleAR, it improves performance from 63.66\% to 71.27\% (+7.61 points), establishing a new state-of-the-art that outperforms all evaluated models, including the 14B-parameter Emu3 (62.31\%), with best performance across most individual benchmarks.


\highlight{CompGen improves compositionality without sacrificing visual quality or overfitting.} 
Figure~\ref{fig:compare} demonstrates CompGen's superior compositional capabilities. 
While strong baselines (SDXL, Lumina-Next) frequently misinterpret object counts (e.g., ``two spring rolls''), fail to bind attributes correctly (e.g., a squirrel ``in a studded leather jacket''), or miss complex relationships (e.g., ``shouting into a microphone''), CompGen accurately renders these details while maintaining visual quality without artifacts. 

Additional examples appear in Appendix~\ref{app:vis}.


\subsection{Impact of Reward Model}
\label{sec:abla_reward}


Here, we investigate the impact of reward model capability on CompGen's performance by comparing four vision-language models as reward models: LLaVA-v1.6-13B~\citep{liu2023visual}, LLaVA-v1.5-13B~\citep{liu2023visual}, CLIP-FlanT5-XXL~\citep{radford2021learning}, and InstructBLIP-FlanT5-XXL~\citep{dai2023instructblip}. 
Table~\ref{tab:reward_impact} shows a strong positive correlation between reward model capability and CompGen performance. 
The strongest model (LLaVA-v1.6-13B) achieves 66.62\% average score, outperforming the weakest (InstructBLIP) by 9.4 percentage points (57.22\%). 
This scaling behavior indicates that CompGen's effectiveness improves directly with vision-language model advancements, pointing to a clear path for future improvements via better reward models.
We offer a visualization in Figure~\ref{fig:abaltioncase_app}.

We also provide additional reward function analysis in Appendix~\ref{app:rewardfunction}. 
Our findings indicate that a fine-grained, multi-aspect reward function provides more effective guidance for complex compositional generation than a coarse-grained reward function.

\subsection{Impact of Difficulty Measures}
\label{sec:abla_difficulty}
We evaluate the effectiveness of different difficulty measures in Table~\ref{tab:difficulty_measure} by training Stable-Diffusion-1.5 on 10K prompts uniformly sampled across difficulty levels 1-10. 
Existing methods \citep{gao2024generate,chen2024makes}  typically employ additive metrics that sum or average compositional components.
In contrast, our proposed difficulty measure uses a multiplicative formulation to better capture the combinatorial explosion in compositional complexity as the number of components increases. 
Results demonstrate that curriculum learning guided by our difficulty measure achieves superior performance across all benchmarks, yielding an average score of 66.62\% — a 4.56 percentage point improvement over the strongest additive baseline. 
This substantial gain underscores the importance of accurately modeling the exponential nature of compositional T2I difficulty.

We further investigate the impact of data difficulty distribution in Appendix~\ref{app:distribution}.
Our results show that a balanced training curriculum covering a wide range of data difficulties is crucial for robust RL training, as training exclusively on skew-easy or skew-hard samples \citep{shi2025efficient} leads to poor compositional generalization.

\begin{figure}
\centering
\includegraphics[width=0.9\linewidth]{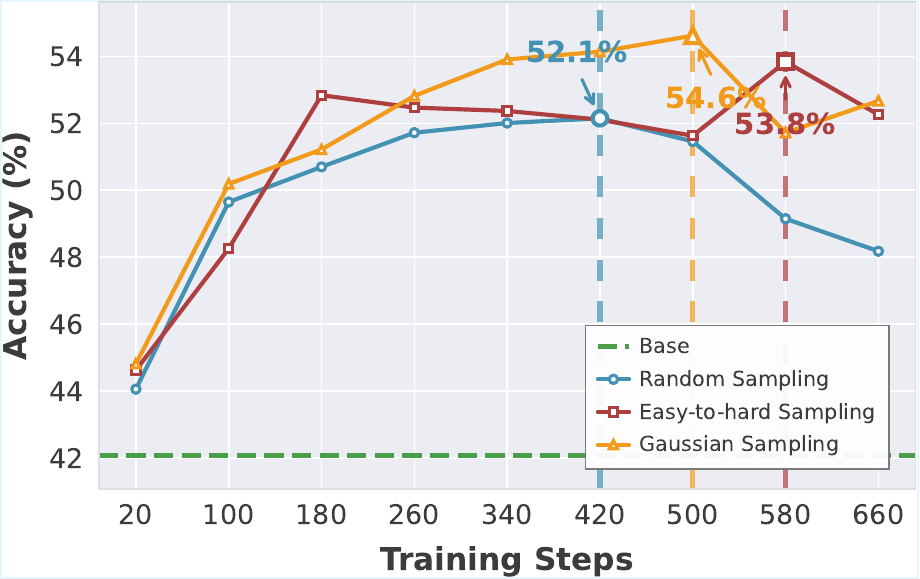}
\vspace{-2mm}
\caption{Scaling trend of CompGen with different curriculum scheduling strategies.}
\label{fig:performance}
\vspace{-2mm}
\end{figure}


\subsection{Analysis of Curriculum Scheduling Strategy}  

\label{sec:abla_curr}

We investigate how curriculum scheduling strategies impact model performance and scaling behavior. 
Using Stable-Diffusion-1.5 on the GenEval~\citep{ghosh2023geneval} benchmark, we compare three scheduling strategies~\citep{parashar2025curriculum}: (i)  \textbf{Random Scheduling}, uniform sampling across all difficulty levels as a baseline; (ii) \textbf{Easy-to-Hard Scheduling}, a deterministic sequential progression from easiest to hardest samples; and (iii) \textbf{Gaussian Scheduling}, a smooth transition where sampling follows a bell-shaped curve that gradually shifts from easy to hard data. 
Models are trained with batch size 32 for up to 660 steps. Further implementation details appear in Appendix~\ref{app:strategy}.
Figure~\ref{fig:performance} demonstrates that curriculum strategies substantially improve CompGen performance. 
While the baseline model without curriculum learning stagnates at 42\% accuracy, all curriculum approaches achieve significant gains. Random sampling provides strong improvement, peaking at 52.1\% accuracy. 
The structured curriculum strategies yield even larger gains: Gaussian scheduling achieves the highest accuracy of 54.6\% at 500 steps — a 30\% relative improvement (12.6 percentage points) over the baseline.
Beyond absolute performance gains, curriculum learning extends the scaling regime compared to conventional training. 
While the baseline plateaus early, curriculum strategies continue improving beyond 500 training steps before gradually declining. 

Notably, Gaussian scheduling demonstrates the most efficient scaling trajectory, reaching optimal performance with fewer steps, while easy-to-hard scheduling exhibits the most extended scaling range, maintaining competitive performance the longest during prolonged training.

\section{Conclusion}
\label{sec:conclusion}
In this work, we introduced CompGen, a novel compositional curriculum reinforcement learning framework that leverages scene graphs and incorporates a principled difficulty criterion with an adaptive Markov Chain Monte Carlo sampling algorithm. 
Our approach free from the requirements of need of ground-truth images, systematically improves the ability of T2I models to generate complex compositional scenes with multiple objects, diverse attributes, and intricate spatial-semantic relationships.
In the future, it would be interesting to explore more sophisticated metrics that incorporate semantic complexity, visual realism requirements, and cross-modal alignment challenges. Additionally, developing adaptive curriculum strategies that dynamically adjust difficulty based on model performance might further optimize training efficiency.

\FloatBarrier

\clearpage

{
    \small
    \bibliographystyle{ieeenat_fullname}
    \bibliography{main}
}

\newpage
\appendix
\clearpage
\setcounter{page}{1}
\maketitlesupplementary

\section{Algorithm}
\label{app:algo}
We provide an overall algorithm of our CompGen in Algorithm~\ref{alg:compgen}.
\begin{algorithm}[h]
\caption{CompGen}
\label{alg:compgen}
\begin{algorithmic}[1]
\Require Pre-trained T2I model with parameters $\theta$, pre-trained MLLM $p_\text{reward}$, curriculum of difficulty ranges $\{[\text{Diff}_{\text{min}}, \text{Diff}_{\text{max}}]_k\}_{k=1}^K$.
\Ensure Fine-tuned T2I model with parameters $\theta'$.

\For{each training step}
    \State \textbf{Phase 1: Curriculum-based Data Synthesis}
    \State Schedule a difficulty range $[\text{Diff}_{\text{min}}, \text{Diff}_{\text{max}}]$ from the curriculum according to Appendix~\ref{app:strategy}.
    \State Generate a scene graph $\mathcal{G}$ (as formulated in Definition~\ref{eqn:difficulty}) via adaptive MCMC, such that $\text{Diff}_{\text{min}} \leq \text{Diff}(\mathcal{G}) \leq \text{Diff}_{\text{max}}$ according to Algorithm~\ref{alg:constrained_sampler}.

    \Statex
    \State \textbf{Phase 2: C-GRPO Optimization for Compositional Generation}
    \State Derive natural language input prompt $T$ from $\mathcal{G}$ using a constrained LLM.
    \State Generate a group of $G$ images $\{I^{(i)}\}_{i=1}^G$ using $T$ as input from policy $p_{\theta_{\text{old}}}(\cdot|T)$.
    \State Generate a set of binary question-answer pairs (e.g., $Q_\text{object}$, $Q_{\text{count}}$, $Q_{\text{attribute}}$, $Q_{\text{relation}}$ with ``Yes'' answers) from $\mathcal{G}$ according to Section~\ref{sec:CGRPO}.
    \State Compute per-question rewards $\widehat{r}_j^{(i)}(t)$ for each image $I^{(i)}$ at step $t$.
    \State Compute overall image reward and normalized advantages $A_i(t)$ at step $t$ according to Eq.~(\ref{eqn:advantage}).
    \State Optimize C-GRPO objective $\mathcal{J}_\text{C-GRPO}(\theta)$ with clipped importance sampling to update $\theta$ using Eq.~(\ref{eqn:loss}).
\EndFor

\State \Return Fine-tuned T2I model with parameters $\theta$.
\end{algorithmic}
\end{algorithm}

\begin{figure*}[t]
    \centering
    \includegraphics[width=\textwidth]{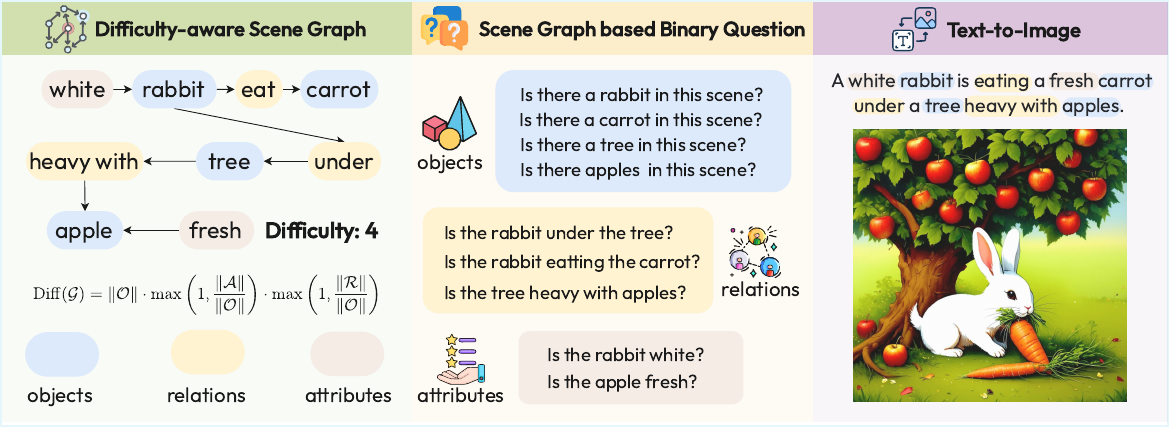}
    \caption{An illustrated example of scene graph complexity (Left), sample binary questions generated from the scene graph (with ``Yes'' answers) (Middle), and corresponding input text and output image pairs (Right).}
    \label{fig:app_difficulty}
    \vspace{-5mm}
\end{figure*}

We also provide an illustrated example of key concepts, including difficulty-aware graphs, scene graph-based binary questions, and text-to-image generation in Figure~\ref{fig:app_difficulty}.


    
        
        
        
    

\begin{algorithm}
\caption{Scene Graph Generation via Adaptive MCMC}
\label{alg:constrained_sampler}
\begin{algorithmic}[1]
    \Require Difficulty bounds $[\text{Diff}_{\text{min}}, \text{Diff}_{\text{max}}]$, max iterations $T$, an annealing schedule for temperature $\tau$
    \Ensure A generated scene graph $\mathcal{G}_T$

    \State Initialize graph $\mathcal{G}_0$ as a simple prior graph
    
    \For{$t \gets 1$ to $T$}
        \State Randomly select a transformation operation $t_{\text{op}} \in \{t_{\text{add}}, t_{\text{delete}}\}$
        \State Propose a candidate graph $\mathcal{G}' \gets t_{\text{op}}(\mathcal{G}_{t-1})$
        
        \State Compute $\text{Energy}(\mathcal{G}_{t-1})$ and $\text{Energy}(\mathcal{G}')$ via Eq.~(\ref{eq:energy})
        \State Decrease temperature $\tau$ according to the annealing schedule
        \State Compute acceptance probability $\text{Acc}(\mathcal{G}'|\mathcal{G}_{t-1})$ via Eq.~(\ref{eqn:acceptance})
        
        \If{$\text{Uniform}(0,1) < \text{Acc}(\mathcal{G}'|\mathcal{G}_{t-1})$}
            \State $\mathcal{G}_t \gets \mathcal{G}'$ \Comment{Accept proposal}
        \Else
            \State $\mathcal{G}_t \gets \mathcal{G}_{t-1}$ \Comment{Reject proposal}
        \EndIf
    \EndFor
    
    \State \Return $\mathcal{G}_T$
\end{algorithmic}
\end{algorithm}

\section{Scene Graph Asset Generation}
\label{app:scenegraph}

To instantiate the sampled scene graphs, we require corresponding data assets. These assets can be categorized into three types: objects, attributes, and relations.
\label{app:asset}
\subsection{Object Generation}
\label{subsec:obj_gen}
To generate a diverse set of object assets that are well-representable by T2I models, we first selected 10 independent and broad object categories~\citep{an2025agfsync}: ``Natural Landscapes'', ``City Infrastructure and Street Elements'', ``People'', ``Animals'', ``Plants'', ``Food and Beverages'', ``Sports and Fitness'', ``Technology Equipment and Industry'', ``Everyday Objects'', and ``Transportation''. Based on these categories, we employed the DeepSeek-V3~\citep{deepseekai2024deepseekv3technicalreport} to generate 50 distinct objects for each, resulting in a total of 500 objects. To avoid duplicate names, we ensured uniqueness through case-normalized matching and assigned incrementing identifiers for traceability. The entire process involved multi-stage validation to ensure structural and semantic consistency, including schema checks on field types, required keys, and cardinality. Invalid outputs were regenerated, while valid ones were persisted atomically to maintain data integrity.


\subsection{Attribute Generation}
Attribute binding is a crucial component of a model's composability capability. To generate attribute assets that correspond to entities, we adopt a hierarchical generation process. Specifically, we first use the DeepSeek-V3~\citep{deepseekai2024deepseekv3technicalreport} to generate attribute concepts corresponding to the object, such as ``color'', ``material'', and ``shape'' for a chair. Next, we employ the DeepSeek-V3~\citep{deepseekai2024deepseekv3technicalreport} to generate valid values for each attribute concept. For example, for the material attribute of a chair, valid values such as ``wood'', ``metal'', and ``plastic'' are generated. Compatibility checks are applied at both stages of the process. For each object, we generate 5 attribute concepts, and for each concept, we generate 5 attribute values.

\label{subsec:attr_gen}

\subsection{Relation Generation}
For relations, our framework adopts a selective generation mechanism that produces semantic relations only for sampled object pairs, avoiding the computational cost of exhaustive pairwise enumeration. Relations are generated exclusively for edge nodes specified in the pre-computed scene graph topology, which ensures structural alignment while enabling semantic enrichment. Each relation is dynamically conditioned on the categorical and attribute-level properties of both participating objects through carefully designed LLM prompts that enforce contextual appropriateness.To maintain uniqueness while supporting diverse interactions across scene configurations, the framework employs a deterministic pair-identification scheme that prevents duplicate object-pair relations. This results in linear computational complexity relative to the number of objects, making the system scalable for complex scenes, while inherently preserving semantic coherence by jointly enforcing structural and attribute-based constraints during generation. Invalid outputs trigger automatic regeneration, whereas validated relations are atomically persisted to guarantee reliability throughout the pipeline.
\label{subsec:rel_gen}

\section{Curriculum Scheduling Strategies}
\label{app:strategy}
\begin{itemize}[left=0pt]
    \item \textbf{Random Scheduling.} 
    As a straightforward method to prevent forgetting, the random scheduling strategy allows for sampling from all data with equal probability throughout the entire training process. This can be viewed as a trivial form of curriculum learning, or more generally, as the default behavior of most policy optimization algorithms when task difficulty is not considered. For $M$ tasks, the sampling probability $p_{\text{random}}(t, j)$ for task $j$ at any training iteration $t$ is set to:

    $$
    p_{\text{random}}(t, j) = \frac{1}{M}
    $$
    This means that at each training step, all tasks have an equal chance of being selected. While this method effectively prevents the model from forgetting previously learned tasks, it might introduce more difficult tasks too early, leading to reward sparsity issues and potentially hindering the curriculum learning strategy from achieving optimal results.
    
    \item \textbf{Easy-to-Hard Scheduling.} Easy-to-Hard curriculum learning strategies adopt a phased approach, dividing the training process into a series of incrementally difficult stages. In each training phase, the model focuses solely on tasks of a specific difficulty level. We can define the sampling indicator function for task $j$ at training step $t$ as $p_{\text{E2H}}(t, j)$. For a total of $M$ tasks ($j=1, \ldots, M$), this function takes a value of 1 if the current training iteration $t$ falls within the predefined stage $[\tau_j, \tau_{j+1}]$ for task $j$; otherwise, it is 0. Here, $\tau_j$ denotes the starting training step for the $j$-th stage, with $\tau_1=0$ and $\tau_{M+1}=N_T$ being the total number of training steps. This implies that at a given iteration $t$, only tasks corresponding to the current stage are sampled for training.

    $$
    p_{\text{E2H}}(t, j) = 
    \begin{cases}
        1, & \text{if } \tau_j \le t < \tau_{j+1} \\
        0, & \text{otherwise}
    \end{cases}
    $$
   Therefore, at training step $t$, the sampling distribution will be 
    $[p_{\text{E2H}}(t, 1), \ldots, p_\text{E2H}(t, M)]$.

     \item \textbf{Gaussian Scheduling.}  
   Gaussian scheduling models task sampling as a mixture of Gaussian distributions to provide flexible and fine-grained control over the training process.

   Each task $j$ ($j=1,\dots,M$) is assumed to follow a one-dimensional Gaussian distribution with the same variance $\sigma^2$ but different means
   \[
   \mu_j = j - 1.
   \]
   A latent curriculum position $x_t$ moves from easier to harder tasks as training step $t$ increases:
   \[
   x_t = \left(\frac{t}{N_T}\right)^{\beta}(M - 1),
   \]
   where $N_T$ is the total number of training steps, $\beta>0$ controls the moving speed of $x_t$, and $M$ is the number of tasks.

   The unnormalized sampling score for task $j$ at step $t$ is
   \[
   s_{\text{Gaussian}}(t,j) =
   \exp\!\left(-\frac{(x_t - \mu_j)^2}{2\sigma^2}\right),
   \]
   where $\sigma$ determines the sampling concentration.

   The normalized sampling probability is
   \[
   p_{\text{Gaussian}}(t,j) =
   \frac{s_{\text{Gaussian}}(t,j)}
   {\displaystyle \sum_{m=1}^M s_{\text{Gaussian}}(t,m)}.
   \]

   Smaller $\sigma$ produces sharper, stage-like transitions, while larger $\sigma$ smooths the task shifts. A lower $\beta$ slows the move toward harder tasks, allowing more training on easy tasks in the early phase.
\end{itemize}

\section{Datasets and Metrics}
\label{app:dataset}
We summarize the benchmarks and crossponding metrics used to evaluate the compositional capabilities of text-to-image models below:
\begin{itemize}[left=0pt]
    \item \textbf{GenEval}~\citep{ghosh2023geneval} is a comprehensive benchmark comprising 553 meticulously designed and highly structured prompts that assess model performance across six key evaluation dimensions: single object, dual objects, color, count, spatial positioning, and attribute binding. The GenEval~\citep{ghosh2023geneval} framework automatically evaluates text-to-image alignment by using object detection to verify object presence, count, and position, and color classification models to verify attributes, calculating binary scores for single objects, dual objects, colors, counts, spatial positioning, and attribute binding, then averaging these scores for an overall compositional quality assessment, we use this overall score as the final score.
    \item \textbf{T2I-CompBench}~\citep{huang2025t2i} is a large-scale benchmark consisting of 6,000 carefully curated compositional text prompts designed to evaluate text-to-image models across varying levels of semantic and structural complexity. The prompts are organized into three primary evaluation domains—attribute binding, object relationships, and complex compositions—and further subdivided into six fine-grained categories: color binding, shape binding, texture binding, spatial relationships, non-spatial relationships, and complex compositions. For attribute binding (color, shape, texture), it employs Disentangled BLIP-VQA~\citep{li2022blip}, which breaks prompts into individual object-attribute questions and calculates the product of the “yes” probabilities. Spatial relationships are verified through UniDet-based object detection using geometric rules such as relative positions and IoU thresholds. Complex compositions combine these with CLIPScore~\citep{hessel2021clipscore} in a 3-in-1 metric that averages the specialized evaluations. We select the complex compositions score as the final score.
    \item \textbf{TIFA}~\citep{hu2023tifa} is a comprehensive evaluation framework designed to assess the generation quality of text-to-image models through a combination of 4,000 diverse, human-curated text prompts and 25,000 automatically generated questions utilizing a Visual Question Answering (VQA) model. The benchmark spans 12 distinct question categories, including existence verification, object count, color identification, and spatial reasoning, each targeting different aspects of visual understanding and semantic alignment. TIFA's~\citep{hu2023tifa} overall score is the percentage of questions that visual models answer correctly when evaluating how well a generated image matches its text description. We use this overall score as the final score.
    \item \textbf{DPG-Bench}~\citep{hu2024ella} is a benchmark that consists of 1,065 densely annotated prompts, each with an average token length of 83.91. These prompts are carefully crafted to describe complex visual scenarios that involve multiple objects, attributes, and modifiers, providing a rich and detailed source for evaluating text-to-image models. The prompts are designed to test a model’s ability to handle intricate compositional structures, where objects are not only depicted in isolation but also in relation to one another, with various contextual modifiers such as color, size, orientation, and spatial relationships. DPG-Bench~\citep{hu2024ella} evaluates models across Global, Entity, Attribute, Relation, and Other categories by using an MLLM judge to score the correctness of answers to scene-graph–based questions for each generated image, and the overall score is simply the average of all prompt-level scores. We use this overall score as the final score.
    \item \textbf{DSG}~\citep{cho2023davidsonian} is an evaluation framework for text-to-image (T2I) generation, designed to address the limitations of the Question Generation and Answering approach. It generates a set of questions from a given prompt, which are then answered by a VQA model. The core components of DSG~\citep{cho2023davidsonian} include questions about entities, their attributes, relationships, and global scene features. Each question is atomic, unique, and semantically complete. These questions are organized into a Directed Acyclic Graph (DAG), where the dependencies between questions are explicitly defined. For example, if a parent question (e.g., whether an object is present) receives a negative answer, its dependent questions (e.g., the object's color) are skipped to prevent inconsistent or irrelevant responses. The overall score is computed by averaging the accuracy of the VQA model’s answers. Each question is validated based on whether its truth value can be clearly and reliably determined, and only such valid questions are included in the evaluation. The system also ensures that no redundant or illogical questions are generated. We use this overall score as the final score.
\end{itemize}

\section{Baseline Models}
\label{app:models}
The baseline models we compare against in our experiments are summarized as follows:

\begin{itemize}[left=0pt]
    \item \textbf{Stable-Diffusion-1.4}~\citep{rombach2022high} is one of the earliest widely-adopted open-source text-to-image latent diffusion models. It was trained on a diverse and large-scale dataset of image-text pairs, which enabled it to generate high-quality images from natural language descriptions. By employing a latent diffusion process, Stable-Diffusion-1.4~\citep{rombach2022high} efficiently navigates high-dimensional image spaces, striking a balance between computational efficiency and image fidelity. This model not only demonstrated the potential of diffusion-based generative models but also established a strong foundation for subsequent versions and related models in the field of generative AI. Its open-source nature has fostered further research and innovation, leading to improvements in model scalability, image resolution, and the diversity of generated outputs. 

    \item \textbf{Stable-Diffusion-1.5}~\cite{rombach2022high} improves upon its predecessor, version 1.4, through refined data curation and additional training iterations, resulting in enhanced performance and greater image generation accuracy. The improved dataset, coupled with extended training cycles, allows the model to better capture intricate details and nuances in both the visual and textual domains. Stable-Diffusion-1.5~\cite{rombach2022high} demonstrated superior robustness in handling more complex prompts, making it a preferred choice for researchers and developers working on advanced generative tasks. Its ability to generate more coherent and contextually appropriate images from intricate or ambiguous text inputs has made it a standard baseline in the field of text-to-image generation, providing a reliable foundation for further model advancements and applications in diverse creative and technical domains.

    \item \textbf{Stable-Diffusion-2.1}~\citep{rombach2022high} enhances high-resolution image synthesis and strengthens semantic alignment between text and image representations. Building upon the architectural and methodological foundations of earlier versions, it introduces refined training pipelines and improved noise scheduling strategies to achieve higher visual fidelity. The model leverages a more carefully curated and filtered dataset, reducing spurious artifacts and improving structural and compositional consistency in generated outputs. As a result, Stable-Diffusion-2.1~\citep{rombach2022high} produces images with sharper details, more accurate object boundaries, and improved correspondence to complex textual descriptions. These advancements make it particularly effective for tasks requiring fine-grained control over visual semantics and have established it as a benchmark for evaluating modern text-to-image diffusion models.

\begin{table*}[tbp]
\vspace{-2mm}
\caption{Investigation on reward function using Stable-Diffusion-1.5 as backbone. Our adopted model is highlighted in \colorbox{TealBlue}{yellow}. The best performance is marked in \textcolor{red}{\textbf{red}}.}
\vspace{-2mm}
\centering
\fontsize{8pt}{10pt}\selectfont
\resizebox{0.9\textwidth}{!}{
\renewcommand{\arraystretch}{1.3}
\begin{tabular}{l ccccccc}
\toprule
\specialrule{0.5pt}{0pt}{0pt}
\rowcolor{lightgray}
\textbf{Model} & \textbf{Reward Model} & \textbf{GenEval} & \textbf{DPG} & \textbf{TIFA} & \textbf{T2I-CompBench} & \textbf{DSG} & \textbf{Avg.} \\
\specialrule{0.5pt}{0pt}{0pt}
\midrule
Stable-Diffusion-1.5 & -- & 42.08\% & 62.24\% & 78.67\% & 29.94\% & 61.57\% & 54.90\% \\
Stable-Diffusion-1.5 w/ VQAScore~\citep{lin2024evaluating} & LLaVA-v1.6-13B & 44.02\% & 73.41\% & 80.19\% & 36.36\% & 73.23\% & 61.44\% \\
\hline
\multirow{4}{*}{CompGen (Stable-Diffusion-1.5)} & CLIP-FlanT5-XXL & 45.11\% & 71.26\% & 81.42\% & 31.60\% & 73.74\% & 60.63\% \\
\cline{2-2}
& InstructBLIP & 42.04\% & 64.00\% & 76.29\% & \best{39.21\%} & 64.58\% & 57.22\% \\
\cline{2-2}
& LLaVA-v1.5-13B & 49.23\% & 74.54\% & 84.84\% & 37.43\% & 75.97\% & 64.40\% \\
\cline{2-2}
& \cellcolor{TealBlue} LLaVA-v1.6-13B & \cellcolor{TealBlue} \best{53.88\%} & \cellcolor{TealBlue} \best{78.67\%} & \cellcolor{TealBlue} \best{85.71\%} & \cellcolor{TealBlue} 37.68\% & \cellcolor{TealBlue} \best{77.16\%} & \cellcolor{TealBlue} \best{66.62\%} \\
\bottomrule
\end{tabular}}
\label{tab:reward_func}
\end{table*}

    \item \textbf{Playground-V2}~\citep{playground-v2} is optimized for creative and vivid image generation, focusing on enhancing aesthetic expressiveness and imaginative scene synthesis. Compared to earlier diffusion-based models, it introduces tuning strategies and architectural refinements that prioritize stylistic diversity, color richness, and artistic coherence. The model demonstrates strong capabilities in generating visually striking compositions that blend realism with creativity, making it particularly suitable for tasks involving conceptual design, digital art, and visual storytelling. By balancing fidelity and artistic abstraction, Playground-V2~\citep{playground-v2} exemplifies how diffusion models can be adapted for open-ended, human-centric creative applications beyond conventional photorealistic synthesis.

    \item \textbf{Stable-Diffusion-XL}~\citep{podell2023sdxl} incorporates a substantially larger model capacity and an improved decoder architecture, leading to significant gains in both visual realism and semantic precision. By expanding the number of parameters and enhancing the expressiveness of its latent representations, the model achieves more detailed, coherent, and photorealistic high-resolution outputs. The redesigned decoder contributes to sharper textures, smoother gradients, and better preservation of global scene structure. In addition, Stable-Diffusion-XL~\citep{podell2023sdxl} demonstrates improved generalization across diverse visual domains, maintaining consistency in complex compositions and fine-grained visual elements. These advancements position it as a state-of-the-art framework for large-scale text-to-image synthesis and a critical reference point for subsequent research in high-fidelity generative modeling.


    \item \textbf{Lumina-Next}~\citep{zhuo2024lumina} represents a new generation of efficient text-to-image models, designed to deliver high-quality image synthesis with significantly reduced inference latency. By optimizing both the model architecture and inference pipelines, Lumina-Next~\citep{zhuo2024lumina} strikes a balance between computational efficiency and visual fidelity, enabling rapid generation of high-resolution images from textual prompts. This improvement in speed, without compromising on quality, makes the model particularly well-suited for interactive applications, where real-time performance is essential. Whether used in creative tools, virtual environments, or user-facing systems, Lumina-Next provides a robust solution for scenarios requiring swift, high-quality visual content generation, while maintaining a responsive and seamless user experience. Its efficiency and scalability position it as a key technology for next-generation interactive AI systems. 

    \item \textbf{LlamaGen}~\citep{sun2024autoregressive} explores a novel language-model-driven paradigm for image generation, bridging the gap between large-scale linguistic understanding and visual synthesis. By integrating powerful language priors from large language models with a dedicated visual decoder, it effectively aligns semantic comprehension with image realization. This design enables the model to interpret nuanced textual descriptions and translate them into coherent and contextually rich visual outputs. LlamaGen~\citep{sun2024autoregressive} demonstrates that linguistic reasoning can significantly enhance visual generation quality, leading to improved semantic consistency, compositional accuracy, and prompt adherence. The approach underscores the emerging synergy between language and vision models, highlighting the potential of unified multimodal architectures for scalable, interpretable, and semantically grounded generative systems. visual decoder, it highlights the synergy between language and vision.  
    \item \textbf{Show-o}~\citep{xie2024show} focuses on complex multi-object scene construction, advancing the capability of diffusion models to generate spatially coherent and semantically aligned compositions. It is designed to handle intricate visual relationships between multiple entities, ensuring that object placement, scale, and interactions conform to natural scene logic. Through enhanced conditioning mechanisms and improved scene representation learning, Show-o~\citep{xie2024show} achieves a higher degree of spatial organization and contextual consistency in generated outputs. The model excels at synthesizing scenes that maintain both local detail fidelity and global structural balance, making it especially valuable for applications in visual reasoning, synthetic dataset generation, and compositional image synthesis. Its emphasis on semantic layout understanding marks an important step toward controllable and interpretable multi-entity generative modeling.

    \item \textbf{SimpleAR}~\citep{wang2025simplear}
    is a lightweight autoregressive (AR) framework for high-quality text-to-image synthesis, focusing on simplicity and effectiveness. It uses a standard next-token prediction approach, where images are tokenized and modeled with text through a unified transformer. With a minimal model of just 0.5B parameters, SimpleAR generates 1024×1024 images with strong fidelity and structure. Its three-stage training pipeline—pretraining, supervised fine-tuning, and reinforcement learning (GRPO \citep{2024DeepSeekMath}) enhances prompt alignment, reasoning, and compositionality. Optimizations like KV-cache acceleration and speculative decoding reduce inference latency, allowing 1024×1024 images to be generated in about 14 seconds. SimpleAR demonstrates that high-quality visual generation can be achieved with a simple, efficient, and scalable AR model.

    \item \textbf{Emu3}~\citep{sun2023emu} is a versatile multi-modal generation system capable of producing not only high-quality images but also videos and other cross-modal outputs. It is architected to enable task generalization across diverse input modalities, including text, audio, and visual signals, thereby supporting a unified generative framework. Through shared latent representations and multi-stage decoding strategies, Emu3 effectively captures complex inter-modal relationships and temporal dependencies, allowing it to synthesize coherent and contextually aligned outputs across formats. This flexibility makes Emu3 particularly suitable for next-generation applications such as interactive media creation, embodied AI, and multi-sensory content generation. Its design reflects a broader trend toward integrated generative systems that seamlessly bridge modalities, pushing the boundary of what multimodal AI models can achieve. 

    \item \textbf{minDALL-E}~\citep{kim2021mindall} is a lightweight variant within the DALL-E family, designed to balance model efficiency with generative capability. Despite its substantially smaller parameter scale, it successfully preserves the core mechanisms underpinning text-to-image synthesis, including semantic understanding and compositional reasoning. The model’s compact architecture enables faster inference and lower computational overhead, making it more accessible for research and educational purposes. By emphasizing reproducibility and open experimentation, minDALL-E~\citep{kim2021mindall} provides a practical platform for studying large-scale generative modeling principles without the extensive resource requirements of its larger counterparts. Its streamlined design illustrates how architectural simplification can coexist with strong generative performance, promoting inclusivity and transparency in multimodal AI research. 
\end{itemize}

\section{Experimental Setup}
\label{app:setup}
\highlight{Training Details of RL.}
For RL training with C-GRPO, we train Stable-Diffusion-1.5 following~\citet{xue2025dancegrpo}'s setting, using a 1e-5 learning rate, AdamW~\citep{loshchilov2017decoupled} optimizer, 1.0 gradient clip norm, a batch size of 32, and generating 12 images per prompt at $512\times512$ resolution with a 1e-4 clip range. For SimpleAR, we adhere to the settings in~\citet{wang2025simplear}, employing a 1e-5 learning rate, AdamW~\citep{loshchilov2017decoupled} optimizer, a batch size of 28, and generating 4 images per prompt at $1024\times1024$ resolution. 
All training is conducted on 8 NVIDIA H800 GPUs.

\begin{figure*}[t]
\begin{center}
\centerline{\includegraphics[width=\textwidth]{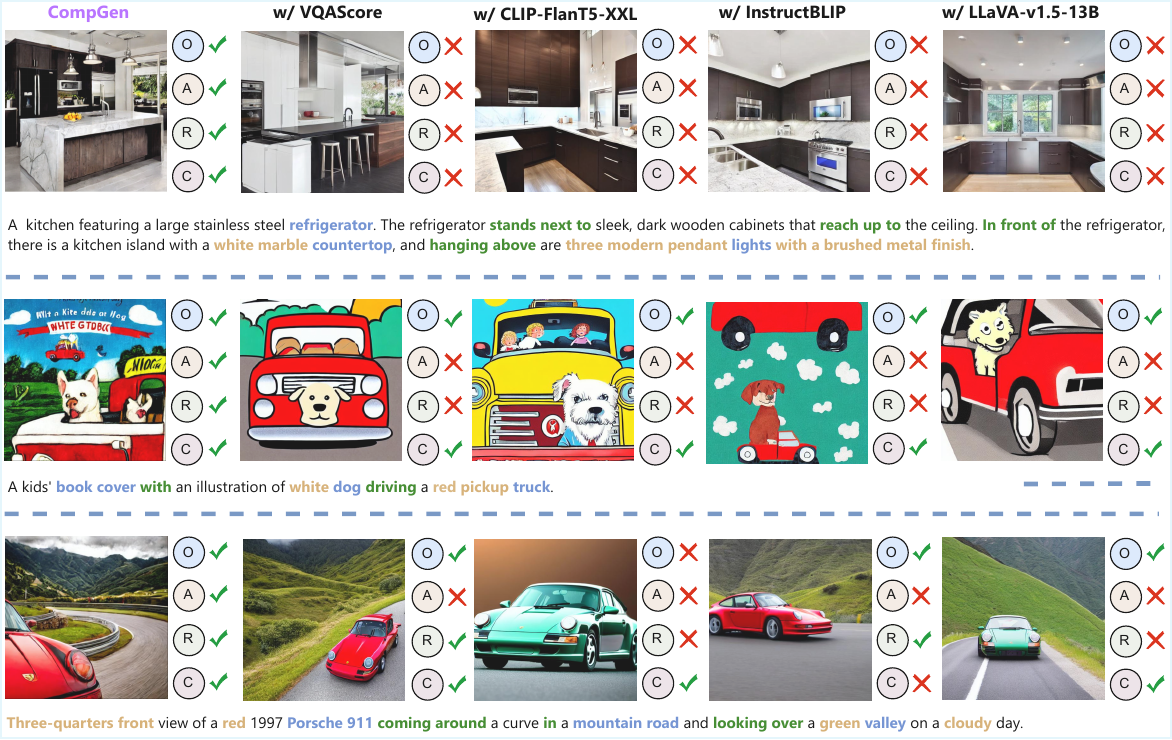}}
\caption{Qualitative comparison of compositional generation across CompGen and ablation models (w/ VQAScore, w/ CLIP-FlanT5-XXL, w/ InstructBLIP, w/ LLaVA-v1.5-13B). Within each prompt, we color the elements for which at least one model makes an error: the object in \objblue{blue}, the attribute in \attbrown{brown}, the relationship in \relgreen{green}, and the count in \coupurple{purple}.
    \raisebox{-0.4ex}{\includegraphics[height=1em]{figures/O_image.pdf}}, 
    \raisebox{-0.4ex}{\includegraphics[height=1em]{figures/A_image.pdf}}, 
    \raisebox{-0.4ex}{\includegraphics[height=1em]{figures/R_image.pdf}}, 
    \raisebox{-0.4ex}{\includegraphics[height=1em]{figures/C_image.pdf}} denote Object, Attribute, Relationship, and Count, respectively. A
    \raisebox{-0.4ex}{\includegraphics[height=1em]{figures/correct_image.pdf}} indicates correct generation, while a   
    \raisebox{-0.4ex}{\includegraphics[height=1em]{figures/wrong_image.pdf}} indicates an error.
}
\label{fig:abaltioncase_app}
\end{center}
\end{figure*}

\section{Additional Ablation Studies}
\label{app:ablation}

\subsection{Impact of Reward Function.}
\label{app:rewardfunction}
To validate our compositional reward function, we conduct an ablation study comparing our fine-grained reward design against the VQAScore baseline~\citep{lin2024evaluating}. While VQAScore uses a single binary question to assess prompt-image alignment, our method decomposes the evaluation into multiple aspects like object existence, attribute binding, relation understanding and numerical counting. As shown in Table~\ref{tab:reward_func}, using the same LLaVA-v1.6-13B reward model \citep{liu2024improved}, our CompGen framework achieves a 66.62\% average score, surpassing VQAScore by a significant 5.2\%. This demonstrates that our fine-grained reward signals more effectively guide the T2I model towards mastering complex compositional generation tasks.

We provide a visual comparison between our method, CompGen, and the VQAScore baseline in Figure~\ref{fig:abaltioncase_app}.

\begin{table}[tbp]
\caption{
Investigation on different \textit{data difficulty distributions} using Stable-Diffusion-1.5 with 10K training data.
The data distribution we adopted is marked with \colorbox{TealBlue}{yellow} background.
The best performance for each benchmark is marked in \textcolor{red}{\textbf{red}}.
``Skew-easy'' means training on easier instances, ``Skew-difficult'' means harder ones, and ``Uniform'' means training on a balanced mix of difficulties..
}
\centering
\resizebox{\columnwidth}{!}{
\fontsize{7pt}{9pt}\selectfont
\renewcommand{\arraystretch}{1.3}
\begin{tabular}{lcccccc}
\toprule
\specialrule{0.5pt}{0pt}{0pt}
\rowcolor{lightgray}
\textbf{Difficulty Distribution} & \textbf{GenEval} & \textbf{DPG} & \textbf{TIFA} & \textbf{T2I-Bench} & \textbf{DSG} & \textbf{Avg.} \\
\specialrule{0.5pt}{0pt}{0pt}
\midrule
Skew-easy 
& 30.24\%
& 40.20\% 
& 48.64\%
& 29.52\%
& 55.64\% 
& 40.85\% \\
Skew-difficult
& 24.78\% 
& 50.88\% 
& 60.31\% 
& 25.68\%
& 50.27\%
& 42.38\% \\
\cellcolor{TealBlue} Ours (Uniform)
& \cellcolor{TealBlue} \textcolor{red}{\textbf{53.88\%}} 
& \cellcolor{TealBlue} \textcolor{red}{\textbf{78.67\%}} 
& \cellcolor{TealBlue} \textcolor{red}{\textbf{85.71\%}} 
& \cellcolor{TealBlue} \textcolor{red}{\textbf{37.68\%}} 
& \cellcolor{TealBlue} \textcolor{red}{\textbf{77.16\%}} 
& \cellcolor{TealBlue} \textcolor{red}{\textbf{66.62\%}} \\
\bottomrule
\end{tabular}}
\label{tab:difficulty_distribution}
\end{table}

\subsection{Impact of Data Difficulty Distribution.}
\label{app:distribution}
We investigate how the training data difficulty distribution affects model performance in Table~\ref{tab:difficulty_distribution}. All experiments are conducted on Stable Diffusion 1.5~\cite{rombach2022high}, with each trained on 10k samples. We compare our uniform data difficulty distribution (uniformly covering difficulty levels 1-10) against two skewed distributions: ``Skew-easy'' (10k samples from levels 1-3) and ``Skew-difficult'' (10k samples from levels 8-10). The results are stark: both skewed distribution perform poorly. Training solely on easy data fails to prepare the model for complex prompts, while focusing only on hard data appears to cause catastrophic forgetting, crippling generalization. Our approach, which provides a balanced curriculum across all difficulty levels, substantially outperforms both, achieving an average score of 66.62\%. This highlights that a curriculum spanning a comprehensive range of difficulties is crucial for enhancing compositional text-to-image generation.

\section{Additional Experiments on Scene Graph Generation}

\begin{table}[tbp]
\caption{
Comparison of sampling efficiency and graph diversity using different sampling methods.
The proposed method is marked with \colorbox{TealBlue}{yellow} background.
The best performance is marked in \textcolor{red}{\textbf{red}}.
\textbf{SR} denotes Success Rate, and \textbf{NTD} denotes Node Type Diversity.
}
\centering
\resizebox{\columnwidth}{!}{
\fontsize{7pt}{9pt}\selectfont
\renewcommand{\arraystretch}{1.3}
\begin{tabular}{lcccccc}
\toprule
\specialrule{0.5pt}{0pt}{0pt}
\rowcolor{lightgray}
& \multicolumn{2}{c}{\textbf{Easy}} & \multicolumn{2}{c}{\textbf{Medium}} & \multicolumn{2}{c}{\textbf{Hard}} \\
\rowcolor{lightgray}
\multirow{-2}{*}{\textbf{Graph Sampling Method}} & \textbf{SR} & \textbf{NTD} & \textbf{SR} & \textbf{NTD} & \textbf{SR} & \textbf{NTD} \\
\specialrule{0.5pt}{0pt}{0pt}
\midrule
Random Rejection Sampling
& 62.50\%
& 39 
& 48.40\%
& 32 
& 35.60\%
& 21 \\
Greedy Sampling
& 90.40\%
& 35 
& 88.30\%
& 46 
& 84.10\%
& 37 \\
\cellcolor{TealBlue} Ours (Adaptive MCMC Sampling)
& \cellcolor{TealBlue} \textcolor{red}{\textbf{98.80\%}} 
& \cellcolor{TealBlue} \textcolor{red}{\textbf{42}} 
& \cellcolor{TealBlue} \textcolor{red}{\textbf{95.20\%}} 
& \cellcolor{TealBlue} \textcolor{red}{\textbf{88}} 
& \cellcolor{TealBlue} \textcolor{red}{\textbf{91.50\%}} 
& \cellcolor{TealBlue} \textcolor{red}{\textbf{130}} \\
\bottomrule
\end{tabular}}
\label{tab:sampling_comparison}
\end{table}

\begin{table}[tbp]
\caption{
Analysis of different initialization strategies for the prior graph.
The default setting used in our approach is marked with \colorbox{TealBlue}{yellow} background.
The best performance is marked in \textcolor{red}{\textbf{red}}.
\textbf{SR} denotes Success Rate, and \textbf{NTD} denotes Node Type Diversity.
}
\centering
\resizebox{\columnwidth}{!}{
\fontsize{7pt}{9pt}\selectfont
\renewcommand{\arraystretch}{1.3}
\begin{tabular}{lcccccc}
\toprule
\specialrule{0.5pt}{0pt}{0pt}
\rowcolor{lightgray}
& \multicolumn{2}{c}{\textbf{Easy}} & \multicolumn{2}{c}{\textbf{Medium}} & \multicolumn{2}{c}{\textbf{Hard}} \\
\rowcolor{lightgray}
\multirow{-2}{*}{\textbf{Initial Strategy of Prior Graph}} & \textbf{SR} & \textbf{NTD} & \textbf{SR} & \textbf{NTD} & \textbf{SR} & \textbf{NTD} \\
\specialrule{0.5pt}{0pt}{0pt}
\midrule
Empty Initial Graph
& 97.40\%
& 37 
& 94.10\%
& 85 
& 87.20\%
& 114 \\
Dense Initial Graph
& 98.00\%
& \textcolor{red}{\textbf{46}} 
& 93.80\%
& 84 
& 89.90\%
& 127 \\
\cellcolor{TealBlue} Ours
& \cellcolor{TealBlue} \textcolor{red}{\textbf{98.80\%}} 
& \cellcolor{TealBlue} 42
& \cellcolor{TealBlue} \textcolor{red}{\textbf{95.20\%}} 
& \cellcolor{TealBlue} \textcolor{red}{\textbf{88}} 
& \cellcolor{TealBlue} \textcolor{red}{\textbf{91.50\%}} 
& \cellcolor{TealBlue} \textcolor{red}{\textbf{130}} \\
\bottomrule
\end{tabular}}
\label{tab:init_robustness}
\end{table}

\subsection{Efficiency of Adaptive MCMC Graph Sampling}
\label{app:mcmc_exp}

To demonstrate the necessity and effectiveness of our proposed Adaptive MCMC sampling approach in navigating the combinatorial graph space, we conduct a comparative study against standard sampling baselines.

\highlight{Baselines.} We employ two baseline strategies for comparison: 
(1) \textit{Random Rejection Sampling}~\citep{blanca2024fast}: To serve as a brute-force lower bound, this method randomly generates a graph from scratch at each step and accepts it only if it strictly satisfies the difficulty constraints.
(2) \textit{Greedy Sampling}~\citep{chamon2017greedy}: To simulate a local optimization approach, this method proposes modifications to the current graph but only accepts transformations that reduce the energy function (i.e., move closer to the target difficulty), unconditionally rejecting any moves that temporarily increase energy.

\highlight{Evaluation Protocol.}
To systematically evaluate performance across varying complexities, we define three difficulty intervals based on the score $\text{Diff}(\mathcal{G})$: \textbf{Easy} ($1 < \text{Diff} \le 4$), \textbf{Medium} ($4 < \text{Diff} \le 7$), and \textbf{Hard} ($7 < \text{Diff} \le 10$). 
For each method and difficulty level, we perform 1,000 independent sampling trials with a maximum budget of $T=100$ iterations per trial.
To quantify the quality of the sampling process, we report two metrics: 
\textbf{Success Rate (SR)}, indicating the percentage of trials that yield a valid graph within the iteration budget; 
and \textbf{Node Type Diversity (NTD)}~\citep{fu2015using}. 
Given that a scene graph comprises object, attribute, and relation nodes, we characterize the graph's structural signature using the tuple $\mathbf{s} = (N_{\text{object}}, N_{\text{attribute}}, N_{\text{relation}})$. NTD quantifies the total number of unique valid signatures $\mathbf{s}$ discovered by the method, serving as a metric for the diversity of the generated graphs.

\highlight{Results Analysis.}
As presented in Table~\ref{tab:sampling_comparison}, \textit{Random Rejection Sampling} exhibits a sharp performance degradation as complexity increases, dropping to a low Success Rate (35.60\%) in the Hard setting. This failure highlights the intractability of blindly searching the high-dimensional graph space.
To improve search efficiency, \textit{Greedy Sampling} achieves respectable success rates in simpler tasks but suffers from mode collapse, evidenced by its significantly lower NTD compared to our method (e.g., 37 vs. 130 in Hard mode). This indicates a tendency to get trapped in local minima, repeatedly reproducing limited structural combinations.
In contrast, to effectively balance exploration and exploitation, our Adaptive MCMC method leverages the Metropolis-Hastings criterion to escape local optima. Consequently, it achieves a Success Rate exceeding 91\% across all levels while maintaining superior semantic diversity (reaching 130 NTD in Hard mode).

\subsection{Performance of Varying Initial Prior Graph}
\label{app:g0}

To verify the robustness of our method, we investigate whether the convergence of the Markov chain is sensitive to the initialization of the prior graph $\mathcal{G}_0$.

\highlight{Baselines.} We compare three distinct initialization strategies:
(1) \textit{Empty Initial Graph}: To test generation from scratch, we initialize with a null graph $\mathcal{G}_0 = \emptyset$.
(2) \textit{Dense Initial Graph}: To test the ability to refine chaotic structures, we initialize with a randomly generated dense graph containing a high number of nodes (4-7) and edges.
(3) \textit{Ours}: To provide a neutral starting point, we initialize with a minimally complex graph containing a small set of random object nodes (1-3).

\highlight{Evaluation Protocol.}
We adhere to the same experimental setup described in Sec.~\ref{app:mcmc_exp}. Specifically, for each initialization strategy, we conduct 1,000 independent trials across the three defined difficulty intervals (Easy, Medium, Hard) with a fixed iteration budget of $T=100$. We evaluate performance using Success Rate (SR) to assess convergence reliability and Node Type Diversity (NTD) to analyze the impact of the starting state on the diversity of the final graph.

\highlight{Results Analysis.}
We analyze the impact of initialization in Table~\ref{tab:init_robustness}. 
First, the overall performance variance across strategies is marginal; for instance, in the Hard setting, the Success Rate remains consistently high for all strategies, confirming the robustness of our Adaptive MCMC sampler to initial states.
Second, regarding the trade-off between efficiency and diversity, the \textit{Dense Initial Graph} yields a slightly higher diversity in the Easy setting (NTD 46 vs. 42). This suggests that starting with high entropy can help explore more combinations in simple tasks, but this complexity results in a lower Success Rate compared to our method (89.90\% vs. 91.50\%) in Hard tasks.
Ultimately, our default initialization strikes the best balance, achieving the highest Success Rate while maintaining competitive or superior graph diversity across all difficulty levels.

\section{Instruction Prompt}
\label{app:prompt}

\subsection{Scene Graph Asset Generation}
\label{app:asset_prompt}
As introduced in Appendix~\ref{app:scenegraph}, scene graph asset generation consists of three components: object generation, attribute generation, and relation generation. 
We provide the corresponding prompt templates in the following prompt boxes.

\newpage
    \onecolumn
    \begin{promptbox}{Prompt Template for Object Generation }
    \label{box:objectgeneration}

\textbf{Task:} Generate exactly 50 unique objects for the category '\texttt{\{category\}}' that are simple, common, and clearly visualizable for use in text-to-image generation.

\medskip

\textbf{Output Format:}

The response must be a JSON dictionary with this exact structure:

\begin{verbatim}
{
    "category_name": "{category}",
    "objects": [
        {"id": 1, "name": "object1"},
        {"id": 2, "name": "object2"},
        ...
    ]
}
\end{verbatim}

\textbf{Requirements:}

\begin{itemize}[leftmargin=*, itemsep=0.2em]
    \item Provide exactly 50 objects
    \item All object names must be unique (case-insensitive comparison)
    \item Object names must be common, easily visualizable examples of the category
    \item Use lowercase names for consistency
    \item Ensure each object can be clearly visualized in a text-to-image setting
    \item Avoid abstract or overly complex objects
    \item Return ONLY the JSON dictionary
\end{itemize}

\medskip

\textbf{Category:} \texttt{\{category\}}

\end{promptbox}
    \begin{promptbox}{Prompt Template for Attribute Generation}
    \label{box:attributegeneration}

\textbf{Task:} Generate a JSON response with exactly 5 VISUALIZABLE attribute concepts and 5 VISUALIZABLE values per concept for the object '\texttt{\{object\_name\}}'.

\medskip

\textbf{Response Structure:}

\begin{verbatim}
{
    "object_name": "{object_name}",
    "attributes": {
        "concept1": ["value1", "value2", ...],
        "concept2": ["value1", "value2", ...],
        ...
    }
}
\end{verbatim}

\textbf{Requirements:}

\begin{itemize}[leftmargin=*, itemsep=0.2em]
    \item Attribute concepts must be:
        \begin{itemize}[nosep]
            \item Instantly visually recognizable and concrete
            \item Directly related to physical appearance or form
            \item Simple, fundamental, not abstract
        \end{itemize}
    \item Values must be:
        \begin{itemize}[nosep]
            \item Single words when possible
            \item Clearly distinct from each other
            \item Common and visually obvious
        \end{itemize}
    \item DO NOT include abstract adjectives like "relaxed", "alert", "matte"
    \item Strictly avoid underscores (\_); use spaces instead
\end{itemize}

\medskip

\textbf{\textcolor{green!60!black}{Good} Example (Simple and highly visual):}

\texttt{\{"object\_name": "painting", "attributes": \{"color": ["red", "blue", "green", "yellow", "black"], "style": ["abstract", "realistic", ...], "size": ["small", "medium", "large", ...]\}\}}

\medskip

\textbf{\textcolor{red}{Bad} Example (Too complex):}

\texttt{\{"color\_palette": ["monochromatic", ...], "brushwork\_style": ["smooth/blended", ...], ...\}}

\medskip

\textbf{Object to analyze:} \texttt{\{object\_name\}}

\end{promptbox}

    \begin{promptbox}{Prompt Template for Relation Generation}
    \label{box:relationgeneration}
\textbf{Task:} Generate one visual relation between two objects in a text-to-image scene.

\textbf{Given objects:}
\begin{itemize}[nosep, leftmargin=*]
    \item Subject: \texttt{\{subject\_name\}}
    \item Object: \texttt{\{object\_name\}}
\end{itemize}

\medskip

\textbf{Requirements:}
\begin{itemize}[nosep, leftmargin=*]
    \item Generate a relation where \texttt{\{subject\_name\}} acts on \texttt{\{object\_name\}}
    \item The relation must represent natural interaction between subject and object
\end{itemize}

\medskip

\textbf{Relation Categories (prioritize spatial):}

\textit{Spatial Relations (PREFERRED):}
\begin{itemize}[nosep, leftmargin=*]
    \item Position: on, under, above, below, beside, next to, near, far from, in front of, behind
    \item Containment: inside, within, outside, around, surrounding, enclosing
    \item Contact: touching, against, leaning on, resting on, attached to, connected to
    \item Orientation: facing, pointing to, directed toward, aligned with
    \item Relative: left of, right of, between, among, across from
\end{itemize}

\textit{Action Relations:}
\begin{itemize}[nosep, leftmargin=*]
    \item Physical: holding, carrying, pushing, pulling, lifting, dropping
    \item Interaction: using, operating, playing with, examining, touching
    \item Movement: approaching, moving toward, following, chasing
\end{itemize}

\textit{Functional Relations:}
\begin{itemize}[nosep, leftmargin=*]
    \item Purpose: for, used by, designed for, intended for
    \item Ownership: belongs to, owned by, part of
    \item State: connected to, linked to, associated with
\end{itemize}

\medskip

\textbf{Examples:}
\begin{itemize}[nosep, leftmargin=*]
    \item person + chair → "sitting on" or "standing beside"
    \item book + table → "on" or "lying on"
    \item car + road → "on" or "driving on"
    \item bird + tree → "perched on" or "flying near"
    \item cup + saucer → "on" or "resting on"
    \item dog + house → "inside" or "in front of"
\end{itemize}

\medskip

\textbf{Output Format:}

Return ONLY the base-form relation word/phrase (no subject/object names, no full sentences).

Examples: "on", "holding", "next to", "inside", "behind"

\end{promptbox}

\subsection{Input Text Generation}
\label{app:input_text_prompt}
We also provide the prompt template to generate the input text in the following prompt box.

\begin{promptbox}{Prompt Template for Input Text Generation}
\label{box:inputtext}

\textbf{Input Format:}

Given the following scene description components in JSON format:

\texttt{\{scene\_graph\}}

\medskip

\textbf{Task:} Generate a single, coherent description that EXACTLY represents all given elements.

\medskip

\textbf{STRICT REQUIREMENTS:}

\begin{itemize}[leftmargin=*, itemsep=0.2em]
    \item ONLY mention relationships EXPLICITLY defined in the "relations" array
    \item DO NOT create new relationships between objects
    \item DO NOT use prepositions like "in", "on", "with", "near" unless they are EXACT relation words in the "relations" array
    \item NEVER group objects together unless they have an explicit relationship
    \item List all objects separately if they don't have relationships with other objects
    \item Include EVERY object with its EXACT attributes as specified
    \item For objects with the same name, use their attributes or explicit relationships to distinguish them
    \item DO NOT use artificial identifiers like "\#1" or "\#2"
    \item MUST use the ORIGINAL wording for ALL elements
    \item MUST NOT add any information not present in the scene graph
    \item List every object instance explicitly. For identical objects, state their count (e.g., 'two identical apples')
\end{itemize}

\medskip

\textbf{EXAMPLES:}

\textit{Scenario 1:} Scene graph: "desert with shrubs", "eagle", "rocket" (NO relations)

\textcolor{red}{BAD:} "In a desert with shrubs, there is an eagle and a rocket"

\textcolor{green!60!black}{GOOD:} "A desert with shrubs, an eagle, and a rocket"

\medskip

\textit{Scenario 2:} Scene graph: "truck", "pizza with olive oil sauce", "truck with a flat cabin type", "motorcycle with a boxy shape", "orange juice", "meadow", "monitor" (NO relations)

\textcolor{red}{BAD:} "There is a truck, a pizza with olive oil sauce, a truck with a flat cabin type..."

\textcolor{green!60!black}{GOOD:} "There are two trucks, one with a flat cabin type, a pizza with olive oil sauce, a motorcycle with a boxy shape, orange juice, a meadow, and a monitor."

\medskip

\textit{Scenario 3:} Scene graph: "dog" (white), "dog" (yellow), "hamburger", relation: "dog (white) eating hamburger"

\textcolor{green!60!black}{GOOD:} "A white dog is eating a hamburger, and a yellow dog is present."

\medskip

\textbf{More Examples:}

\textcolor{red}{Bad:} "A fire hydrant, a slim router, an adult, a waterfall, a cow, a manhole cover, a waterfall with a veil shape, and another manhole cover."

\textcolor{green!60!black}{Good:} "A fire hydrant, a slim router, an adult, two distinct waterfalls (one standard and one veil-shaped), a cow, and two standard manhole covers."

\medskip

\textbf{Output Format:}

Output ONLY this JSON format with NO additional text:

\texttt{\{\{"prompt": "Generated description here"\}\}}

\end{promptbox}

\section{Additional Visualization Results}
\label{app:vis}
\begin{figure*}[t]
    \centering
    \vspace{-9mm}
    \includegraphics[width=\textwidth]{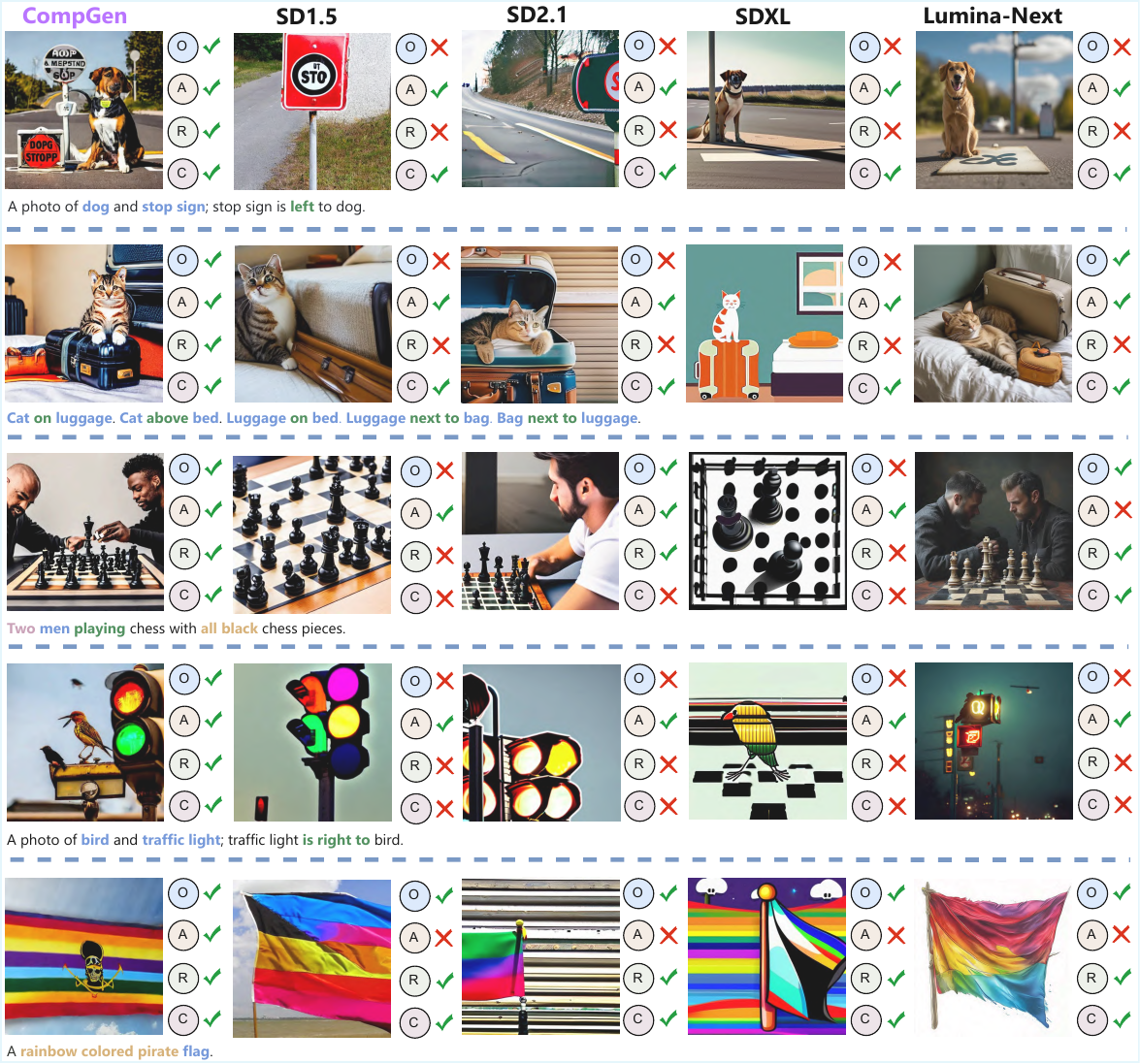}
    \vspace{-7mm}
    \caption{Extended qualitative comparison of our CompGen with other strong text-to-image models (SD1.5, SD2.1, SDXL, and Lumina-Next). Within each prompt, we color the elements for which at least one model makes an error: the object in \objblue{blue}, the attribute in \attbrown{brown}, the relationship in \relgreen{green}, and the count in \coupurple{purple}.
    \raisebox{-0.4ex}{\includegraphics[height=1em]{figures/O_image.pdf}}, 
    \raisebox{-0.4ex}{\includegraphics[height=1em]{figures/A_image.pdf}}, 
    \raisebox{-0.4ex}{\includegraphics[height=1em]{figures/R_image.pdf}}, 
    \raisebox{-0.4ex}{\includegraphics[height=1em]{figures/C_image.pdf}} denote Object, Attribute, Relationship, and Count, respectively. A
    \raisebox{-0.4ex}{\includegraphics[height=1em]{figures/correct_image.pdf}} indicates correct generation, while a   
    \raisebox{-0.4ex}{\includegraphics[height=1em]{figures/wrong_image.pdf}} indicates an error.}
    \vspace{-2mm}
    \label{fig:caseapp1}
\end{figure*}

\begin{figure*}[t]
    \centering
    \vspace{-9mm}
    \includegraphics[width=\textwidth]{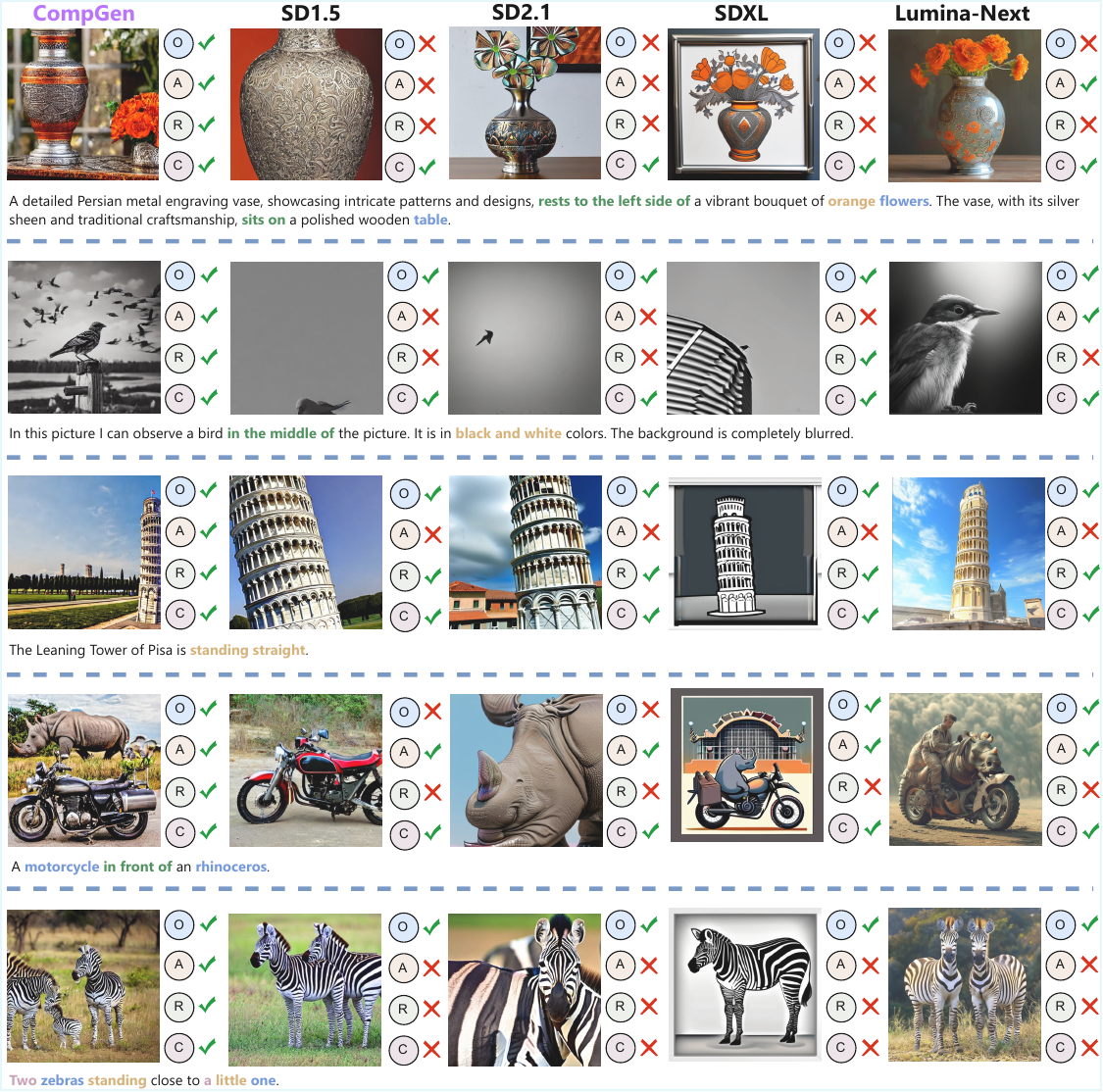}
    \vspace{-7mm}
    \caption{Extended qualitative comparison of our CompGen with other strong text-to-image models (SD1.5, SD2.1, SDXL, and Lumina-Next). Within each prompt, we color the elements for which at least one model makes an error: the object in \objblue{blue}, the attribute in \attbrown{brown}, the relationship in \relgreen{green}, and the count in \coupurple{purple}.
    \raisebox{-0.4ex}{\includegraphics[height=1em]{figures/O_image.pdf}}, 
    \raisebox{-0.4ex}{\includegraphics[height=1em]{figures/A_image.pdf}}, 
    \raisebox{-0.4ex}{\includegraphics[height=1em]{figures/R_image.pdf}}, 
    \raisebox{-0.4ex}{\includegraphics[height=1em]{figures/C_image.pdf}} denote Object, Attribute, Relationship, and Count, respectively. A
    \raisebox{-0.4ex}{\includegraphics[height=1em]{figures/correct_image.pdf}} indicates correct generation, while a   
    \raisebox{-0.4ex}{\includegraphics[height=1em]{figures/wrong_image.pdf}} indicates an error.}
    \vspace{-2mm}
    \label{fig:caseapp2}
\end{figure*}

In Section~\ref{sec:performance}, our visualized examples in Figure~\ref{fig:compare} demonstrate CompGen's superior compositional capabilities over strong baselines (SDXL, Lumina-Next). 
Here, we provide additional visualized results in Figures~\ref{fig:caseapp1} and \ref{fig:caseapp2}.

\end{document}